\documentclass[acmtog]{acmart}

\copyrightyear{2026}
\acmYear{2026}
\setcopyright{cc}
\setcctype{by}
\acmConference[SIGGRAPH Conference Papers '26]{Special Interest Group on Computer Graphics and Interactive Techniques Conference Conference Papers}{July 19--23, 2026}{Los Angeles, CA, USA}
\acmBooktitle{Special Interest Group on Computer Graphics and Interactive Techniques Conference Conference Papers (SIGGRAPH Conference Papers '26), July 19--23, 2026, Los Angeles, CA, USA}
\acmDOI{10.1145/3799902.3811081}
\acmISBN{979-8-4007-2554-8/2026/07}

\usepackage[utf8]{inputenc}
\usepackage{pgfplots}
\usepackage{booktabs} 
\usepackage{multirow} 

\usepackage[ruled]{algorithm2e} 

\SetAlFnt{\small}
\SetAlCapFnt{\small}
\SetAlCapNameFnt{\small}
\SetAlCapHSkip{0pt}

\usepackage{ulem} 
\usepackage{xcolor}
\usepackage{pifont}

\newcommand{\rev}[1]{{\color{black}#1}}

\definecolor{myImageBlue}{RGB}{15,158,213} 
\definecolor{myImageOrange}{RGB}{233,113,50} 

\definecolor{blue}{rgb}{0.290,0.565,0.886}      
\definecolor{sky}{rgb}{0.490,0.725,0.965}       
\definecolor{cyan}{rgb}{0.235,0.737,0.765}      
\definecolor{teal}{rgb}{0.165,0.682,0.620}      
\definecolor{green}{rgb}{0.298,0.722,0.439}     
\definecolor{lime}{rgb}{0.647,0.839,0.290}      
\definecolor{yellow}{rgb}{0.949,0.788,0.298}    
\definecolor{amber}{rgb}{0.961,0.651,0.137}     
\definecolor{orange}{rgb}{0.941,0.541,0.294}    
\definecolor{coral}{rgb}{0.910,0.427,0.463}     
\definecolor{red}{rgb}{0.878,0.333,0.380}       
\definecolor{magenta}{rgb}{0.769,0.353,0.682}   
\definecolor{purple}{rgb}{0.557,0.404,0.816}    
\definecolor{darkgray}{rgb}{0.294,0.333,0.388}  
\definecolor{gray}{rgb}{0.604,0.627,0.651}      
\definecolor{lightgray}{rgb}{0.898,0.906,0.922} 
\definecolor{sand}{rgb}{0.906,0.843,0.694}      

\newcommand{\eg}{e.g.}

\begin{document}

\title{PointLLM-R: Enhancing 3D Point Cloud Reasoning via Chain-of-Thought}

\author{Chaoqi Chen}
\email{cqchen1994@gmail.com}
\authornote{Equal contribution}
\affiliation{%
     \department{Visual Computing Research Center (VCC), College of Computer Science and Software Engineering (CSSE)}
	\institution{Shenzhen University}
	\country{China}	
}

\author{Qile Xu}
\email{xql438814395@gmail.com}
\authornotemark[1]
\affiliation{%
   \department{VCC, CSSE}
	\institution{Shenzhen University}
	\country{China}	
}

\author{Wenjun Zhou}
\email{wenjun.9707@gmail.com}
\affiliation{%
    \department{VCC, CSSE}
	\institution{Shenzhen University}
	\country{China}	
}

\author{Hui Huang}
\email{hhzhiyan@gmail.com}
\authornote{Corresponding author: Hui Huang (hhzhiyan@gmail.com)}
\affiliation{%
    \department{VCC, CSSE}
	\institution{Shenzhen University}
	\country{China}	
}



\begin{abstract}

Understanding 3D point clouds through language remains a fundamental challenge in computer graphics and visual computing, due to the irregular structure of point cloud data and the lack of explicit reasoning in existing 3D multimodal models.
While Chain-of-Thought (CoT) reasoning has shown strong effectiveness in LLMs and image-based MLLMs, its extension to 3D understanding remains largely underexplored.
In this paper, we propose a data-centric framework for constructing large-scale CoT supervision tailored to 3D point cloud understanding.
Our framework consists of a two-stage pipeline that first refines point-text instruction data via vision-language-model-based quality evaluation and reference-guided refinement, and then synthesizes high-quality reasoning paths through Human-in-the-Loop Prompt Optimization (HiLPO).
Using this approach, we build PoCoTI, a CoT-enhanced point-text instruction-following dataset containing 55K samples with explicit reasoning paths.
Fine-tuning PointLLM on PoCoTI yields PointLLM-R, a reasoning-capable 3D multimodal language model.
Extensive experiments on generative 3D classification and captioning demonstrate that PointLLM-R achieves state-of-the-art performance and generalizes robustly to real-world scanned point clouds and multi-turn dialogue scenarios.

\end{abstract}



\begin{teaserfigure}
    \begin{center}
    \centering
    \includegraphics[width=1\textwidth]{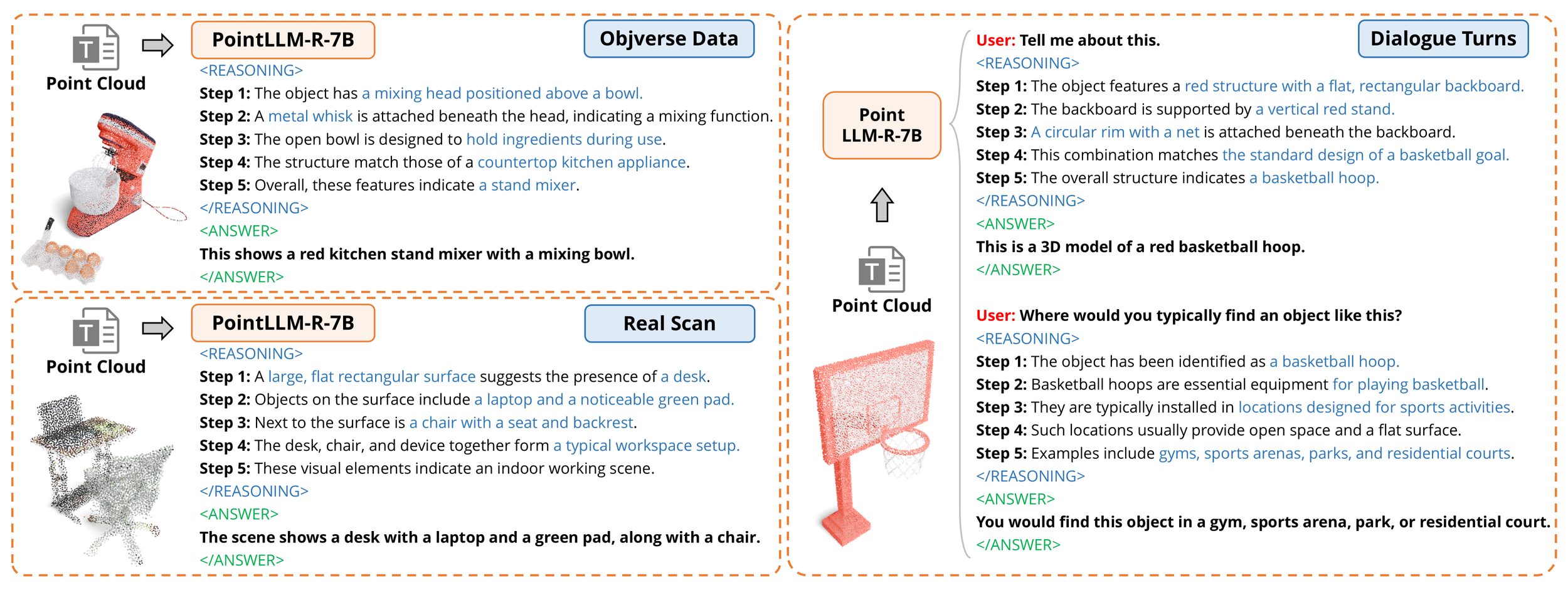}
    \vspace{-8mm}
    \caption{
    PointLLM-R leverages explicit Chain-of-Thought reasoning to enable robust 3D understanding across diverse scenarios. 
   Top left: On dataset samples (Objaverse), it produces structured descriptions grounded in spatial and semantic cues. Bottom left: On real-world scans, it accurately identifies fine-grained object details. Right: In multi-turn dialogues, it maintains coherent reasoning for context-aware question answering over point clouds.
    }
    \label{fig:teaser}
    \end{center}
\end{teaserfigure}


\maketitle

\section{Introduction}
Understanding 3D point clouds is a fundamental problem in computer graphics, with wide-ranging applications in robotics, embodied AI, and AR/VR.
However, point clouds are sparse, unordered, and often incomplete observations of underlying geometry, which makes semantic understanding and geometric reasoning particularly challenging.
Moreover, large-scale annotation of 3D data is expensive and difficult, significantly hindering the development of generalizable 3D understanding models.

Recent progress in multimodal learning has enabled large language models to incorporate visual information and perform open-ended understanding across modalities.
In the 2D domain, models such as Qwen-VL series~\cite{bai2025qwen2.5vl, bai2025qwen3vl} demonstrate strong capabilities by combining vision encoders with large language models.
Motivated by these advances, several works have explored extending multimodal language models to the 3D domain by aligning point cloud representations with language models~\cite{qi2024shapellm, xu2024pointllm, tang2024minigpt}, enabling tasks such as generative classification and captioning directly from raw point clouds.

Despite these advances, existing 3D multimodal models often produce single-step predictions without explicit reasoning.
Such behavior limits interpretability and robustness, especially for queries requiring multi-step inference or fine-grained geometric reasoning.
In contrast, Chain-of-Thought (CoT) reasoning has proven effective in improving reasoning performance and generalization in language models and 2D multimodal models by encouraging step-by-step inference~\cite{wei2022chain, zheng2023ddcot}.
However, enabling CoT reasoning for 3D point cloud understanding remains challenging, largely due to the lack of high-quality 3D datasets with explicit, geometrically grounded reasoning annotations.

To address this limitation, we propose a data-centric framework for constructing CoT supervision tailored to 3D point cloud understanding.
Our approach adopts a two-stage CoT data generation pipeline.
The first stage refines an initial point-text dataset through VLM-based quality evaluation and reference-guided refinement to ensure semantic relevance and factual consistency.
The second stage introduces \textit{Human-in-the-Loop Prompt Optimization (HiLPO)}, which iteratively improves a structured CoT generation prompt via VLM generation, LLM-based refinement, and human verification.
Through this process, we construct \textit{PoCoTI}, a large-scale CoT-enhanced point-text instruction dataset containing approximately 55K samples with explicit reasoning paths.

Built upon PoCoTI, we present \textit{PointLLM-R}, a 3D multimodal large language model endowed with explicit reasoning ability over point cloud inputs.
By fine-tuning PointLLM on PoCoTI, PointLLM-R learns to reason over geometric cues before producing final answers, resulting in more accurate, interpretable, and robust predictions across diverse 3D understanding tasks.

We conduct extensive experiments on generative 3D object classification and 3D object captioning benchmarks, including ModelNet40, Objaverse, and Cap3D.
Importantly, we further evaluate our model on real-world scanned point clouds from OmniObject3D, which exhibit significant geometric noise and domain shift.
Experimental results demonstrate that PointLLM-R consistently outperforms prior 3D multimodal models and generalizes well to challenging real-scanned data, validating the effectiveness of our CoT-enhanced data generation framework.

In summary, our contributions are three-fold:
\begin{itemize}
    \item We propose a two-stage pipeline for constructing geometrically grounded, high-quality CoT supervision for 3D point clouds, combining data refinement and Human-in-the-Loop Prompt Optimization.
    \item We build \textit{PoCoTI}, a large-scale CoT-enhanced point-text instruction dataset with approximately 55K samples and explicit reasoning paths.
    \item We present \textit{PointLLM-R}, a reasoning-capable 3D multimodal large language model that achieves state-of-the-art performance on generative 3D understanding tasks and generalizes well to real-scanned point cloud data.
\end{itemize}

\section{Related Work}

\begin{figure}[ht]
    \centering
    \includegraphics[width=\linewidth]{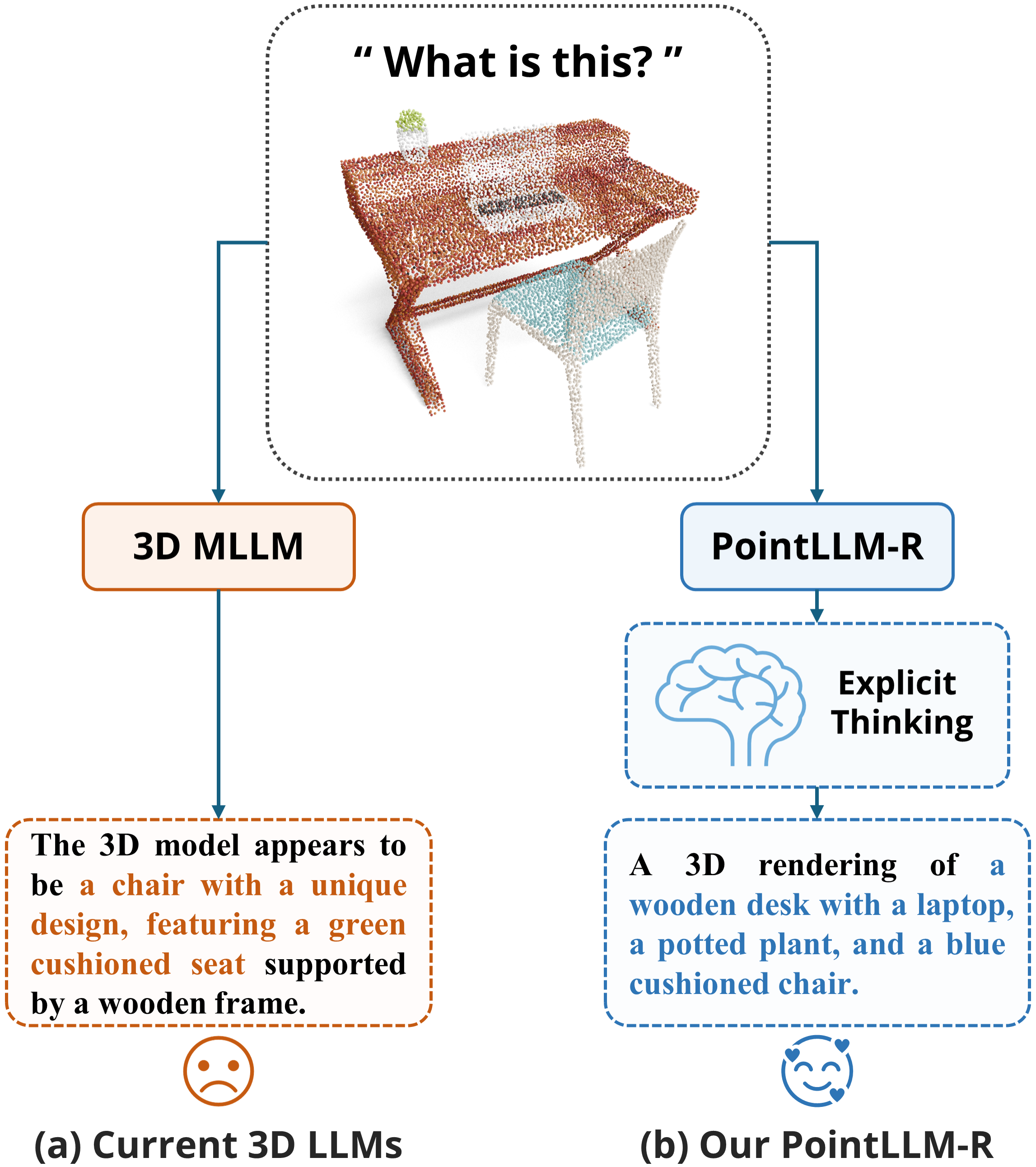}
    \vspace{-3mm}
    \caption{
    Comparison of 3D point cloud understanding behaviors. Existing 3D MLLMs often produce incomplete outputs due to limited reasoning, whereas PointLLM-R employs explicit CoT reasoning for richer and more accurate semantic interpretation.
    }
    \label{fig:method-compare}
    \vspace{-3mm}
\end{figure}

\begin{figure*}[t]
    \begin{center}
    \centering
    \includegraphics[width=1\textwidth]{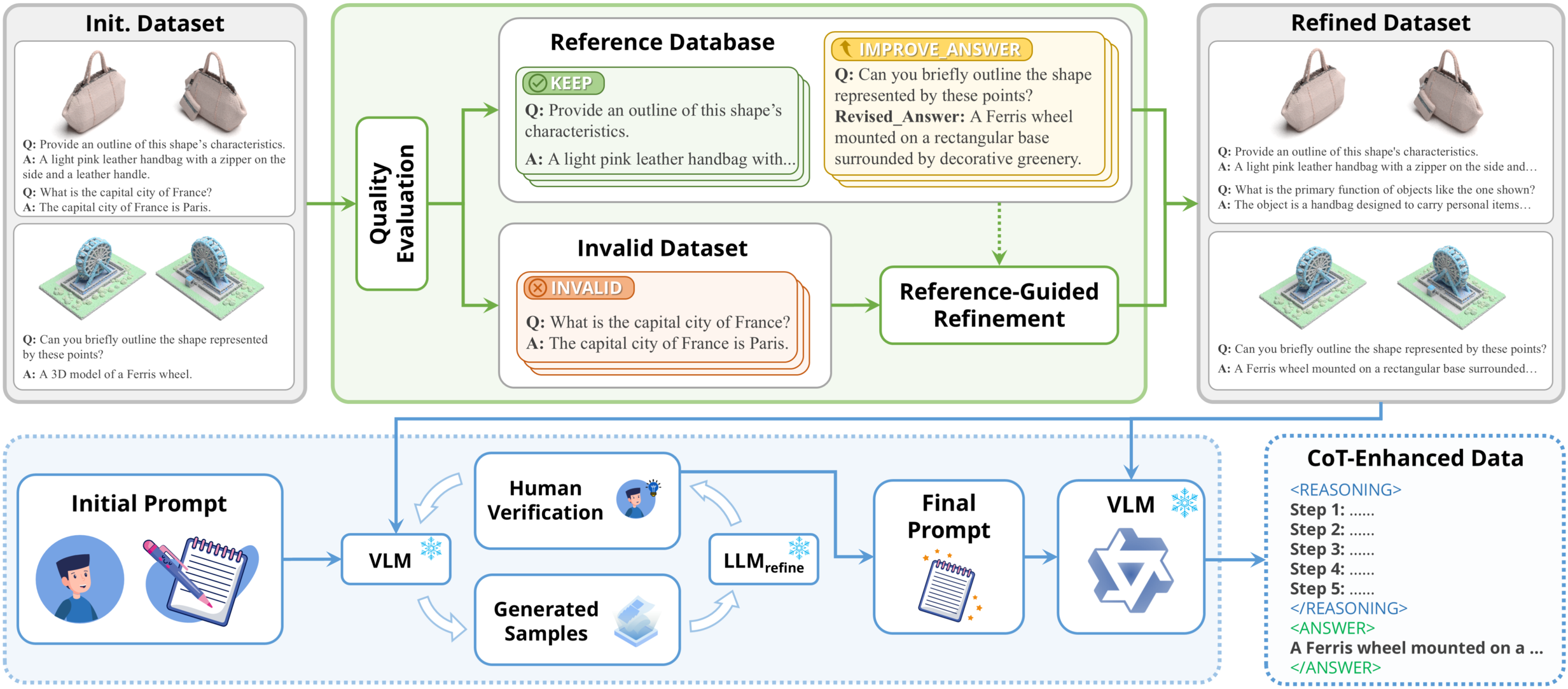}
    \vspace{-7mm}
    \caption{Overview of our two-stage CoT data generation pipeline.
    The first stage refines an initial point-text dataset via VLM-based quality evaluation and reference-guided refinement.
    The second stage adopts Human-in-the-Loop Prompt Optimization (HiLPO) to iteratively optimize a structured CoT generation prompt through VLM generation, LLM-based prompt refinement, and human verification, enabling scalable synthesis of CoT-enhanced data.}
    \label{fig:overview}
    \end{center}
    \vspace{-3mm}
\end{figure*}

\paragraph{Multimodal Large Language Models.}
In recent years, Large Language Models (LLMs) have achieved remarkable progress in natural language processing~\cite{guo2025deepseek, achiam2023gpt, brown2020language, touvron2023llama, team2023gemini}, while Large Vision Models (LVMs) demonstrate strong perceptual capabilities but limited reasoning ability~\cite{shen2024aligning, kirillov2023segment, zhang2022dino, oquab2023dinov2}.
Building upon these advances, image-based MLLMs align visual and textual representations to enable open-vocabulary visual understanding~\cite{bai2025qwen2.5vl, li2025lrm, yang2025storyllava, yan2025tg, li2023blip2}, with representative models including LLaVa~\cite{liu2023visual, liu2024improved} and GPT-4V~\cite{achiam2023gpt}.
Following this paradigm, MLLMs have been extended to other modalities such as video~\cite{zhang2023video, li2024llama, chen2024videollm, tang2025empowering} and audio~\cite{ma2025language, huang2024audiogpt}.
Prior to the adoption of MLLMs in 3D vision, point cloud understanding mainly relied on task-specific neural architectures~\cite{qi2017pointnet, qi2017pointnet++, wang2019dynamic, maturana2015voxnet, wang2017cnn}.
Recently, a growing body of work has explored aligning 3D point cloud representations with language models~\cite{qi2024shapellm, xu2024pointllm, yu2022point, hong20233d, tang2024minigpt}, enabling multimodal understanding of 3D objects.
However, as illustrated in Fig.~\ref{fig:method-compare}, existing 3D MLLMs often lack explicit intermediate reasoning and thus struggle to produce holistic and interpretable outputs.
In contrast, our work focuses on instilling explicit CoT reasoning for 3D point cloud understanding.


\paragraph{Multimodal Chain-of-Thought Reasoning.}
Recent advances have explored in-context learning~\cite{brown2020language, gao2025aim, cahyawijaya2024llms} and CoT reasoning~\cite{wei2022chain, ji2024chain} to enhance complex reasoning in LLMs. 
Multimodal CoT (MCoT) extends this paradigm to multiple modalities, enabling structured reasoning over multimodal inputs.
Driven by large-scale image-based tasks, MCoT has been widely adopted in Visual Question Answering~\cite{wang2024cog, gao2024cantor, shao2024visual, mondal2024kam, wu2024dettoolchain}, with structured reasoning mechanisms proposed to improve interpretability~\cite{hu2025socratic}.
Beyond static images, MCoT has also been introduced into video-based MLLMs to support temporal and dynamic multimodal reasoning~\cite{li2023intentqa, wang2024videocot, fei2024video, hu2025cos}.
\rev{Some recent works begin to explore multimodal reasoning in  scene-level settings, such as video-grounded 3D reasoning~\cite{linghu2026scenecot, tang2025lego, yuan2025scene}. 
Several works further demonstrate MCoT’s effectiveness in 3D generation~\cite{yuan20243d, katara2024gen2sim, yamada2025l3go, wang2025chat2layout}.
However, explicit and structured CoT reasoning for 3D object understanding remains limited, largely due to the lack of high-quality 3D datasets with CoT annotations.}

\section{Proposed Method}

To endow 3D MLLMs with CoT reasoning capability, we propose a \textit{two-stage CoT data generation} framework.
First, we introduce a \textit{Data Refinement} pipeline (Sec.~\ref{sec:refinement}) to clean and enhance point-text instructions, establishing a reliable semantic foundation.
Second, we propose \textit{Human-in-the-Loop Prompt Optimization (HiLPO)} (Sec.~\ref{sec:HiLPO}) to iteratively optimize the reasoning prompt and synthesize the large-scale \textit{PoCoTI} dataset.
Built upon this reasoning-enriched corpus, we finally present \textit{PointLLM-R} (Sec.~\ref{sec:PointLLM-R}), a 3D MLLM capable of complex sequential reasoning.

\vspace{-3mm}
\subsection{Initial Data Collection and Refinement}
\label{sec:refinement}

\paragraph{Initial Data Collection.}
We collect an initial dataset $\mathcal{D}_{\text{init}}$ consisting of approximately 55K $(P, I, A)$ triplets, where $P$ denotes the 3D point cloud, $I$ is the textual instruction, and $A$ is the corresponding answer.
The majority ($\sim$45K) are sourced from the ShapeLLM Supervised Fine-Tuning (SFT) dataset $D_{\text{ShapeLLM}}$~\cite{qi2024shapellm}, contributing complex semantic and functional instructions.
To further diversify the descriptive content, an additional $\sim$10K triplets are formulated by aligning unique point clouds from $D_{\text{ShapeLLM}}$ with their Cap3D~\cite{luo2023scalable} captions.
For these samples, the caption is treated as the answer $A$ responding to a standardized instruction $I$ (\eg, ``Describe this object.''). In this initial dataset, each unique point cloud $P$ is associated with multiple distinct $(I, A)$ pairs.

\vspace{-3mm}
\paragraph{Quality Evaluation.}
Despite the broad foundation of $\mathcal{D}_{\text{init}}$, we observe two critical defects: semantic irrelevance and low-quality responses.
Specifically, semantic irrelevance occurs when questions are decoupled from the visual context, such as inquiring about general geographical facts. Meanwhile, low-quality responses appear as factual hallucinations that contradict the geometric reality or insufficient descriptions that lack visual details.

To address these issues, we designed a pipeline for data quality evaluation and refinement, illustrated in Fig.~\ref{fig:overview}. 
The evaluation revolves around three core dimensions: (i) Question Relevance: whether the instruction $I$ is grounded in observable geometric or semantic attributes of the object; (ii) Answer Accuracy: whether the response $A$ is factually consistent with the object; (iii) Answer Completeness: whether the response provides sufficient descriptive detail to support subsequent reasoning.

Specifically, we rendered each point cloud into four distinct views:
\begin{equation}
    V_{P} = f_{\text{render}}(P) = \{v_1, v_2, v_3, v_4\}.
    \label{eq:rendering}
\end{equation}
where $v_i$ denotes the image from the $i$-th viewpoint.Then we used Qwen3-VL\footnote{In this paper, Qwen3-VL refers to Alibaba's "Qwen3-VL-8B-Instruct"}~\cite{bai2025qwen3vl} as a quality evaluator to assess each sample $(V_{P}, I, A)$.
The evaluator is guided by a structured prompt that incorporates role definition, task decomposition, and decision rules.
For each $(V_{P}, I, A)$ sample, it outputs a classification label $C \in \{ \texttt{KEEP}, \texttt{IMPROVE}, \texttt{INVALID} \}$ and detailed reasons for the decision. 
$\texttt{KEEP}$ denotes high-quality samples to be retained;
$\texttt{IMPROVE}$ indicates valid questions with answers that contain factual hallucinations or lack sufficient descriptive detail;
and $\texttt{INVALID}$ is assigned when the question is irrelevant to the visual context or logically unsound.
For samples categorized as $\texttt{IMPROVE}$, the evaluator also provides a refined response $A'$ for subsequent refinement.

\paragraph{Reference-Guided Refinement.}
Based on the evaluation outcomes, we execute a stratified post-processing pipeline to maximize initial data utilization.
First, samples labeled as \texttt{KEEP} and \texttt{IMPROVE} are accepted into a reference database, while the original answers $A$ of the latter are updated with the refined versions $A'$, establishing a verified semantic ground truth. 
Subsequently, for samples categorized as \texttt{INVALID}, instead of simply discarding them, we leverage the reference database to guide a refinement process. 
Specifically, for each sample, we retrieve the valid $(I, A)$ pairs from the reference database associated with the same point cloud $P$. 
Conditioned on these reference $(I, A)$ pairs and the rendered views $V_P$, we prompt the aforementioned evaluator to first re-evaluate the original question within this enriched context.
If the question is deemed reasonable, the evaluator refines its answer;
otherwise, the evaluator synthesizes a completely new $(I, A)$ pair.
To ensure data diversity, we explicitly instruct the evaluator to generate content distinct from the retrieved references. 
In this way, the reference database serves a dual purpose: providing factual background to ground the re-evaluation, while acting as a comparison set to ensure data diversity during re-generation.
Upon completion of this pipeline, we aggregate the retained, refined, and regenerated samples to form our refined dataset, denoted as $\mathcal{D}_{\text{refined}}$. This corpus ensures semantic alignment and factual consistency, setting a solid foundation for the subsequent reasoning generation phase.

\begin{figure}[t]
    \begin{center}
    \centering
    \includegraphics[width=0.47\textwidth]{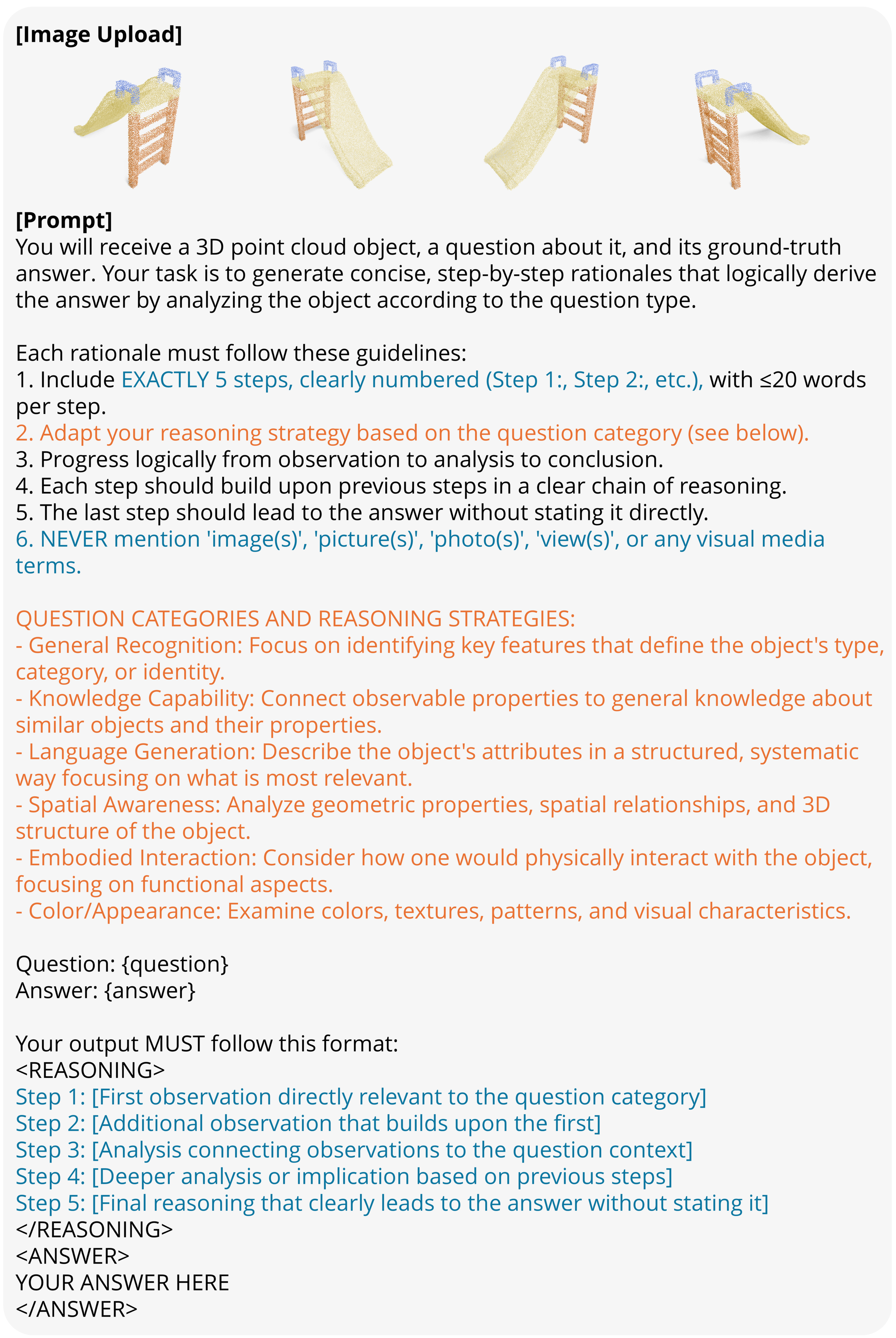}
    \vspace{-3mm}
    \caption{
    The prompt $P^*$ utilized for CoT-Enhanced data generation. Refinements introduced after the first HiLPO iteration are highlighted in \textcolor{myImageBlue}{blue}, while those from the second iteration are highlighted in \textcolor{myImageOrange}{orange}.
    }
    \label{fig:CoT_Prompt}
    \end{center}
    \vspace{-5mm}
\end{figure}

\begin{figure}[t]
    \begin{center}
    \centering
    \includegraphics[width=0.47\textwidth]{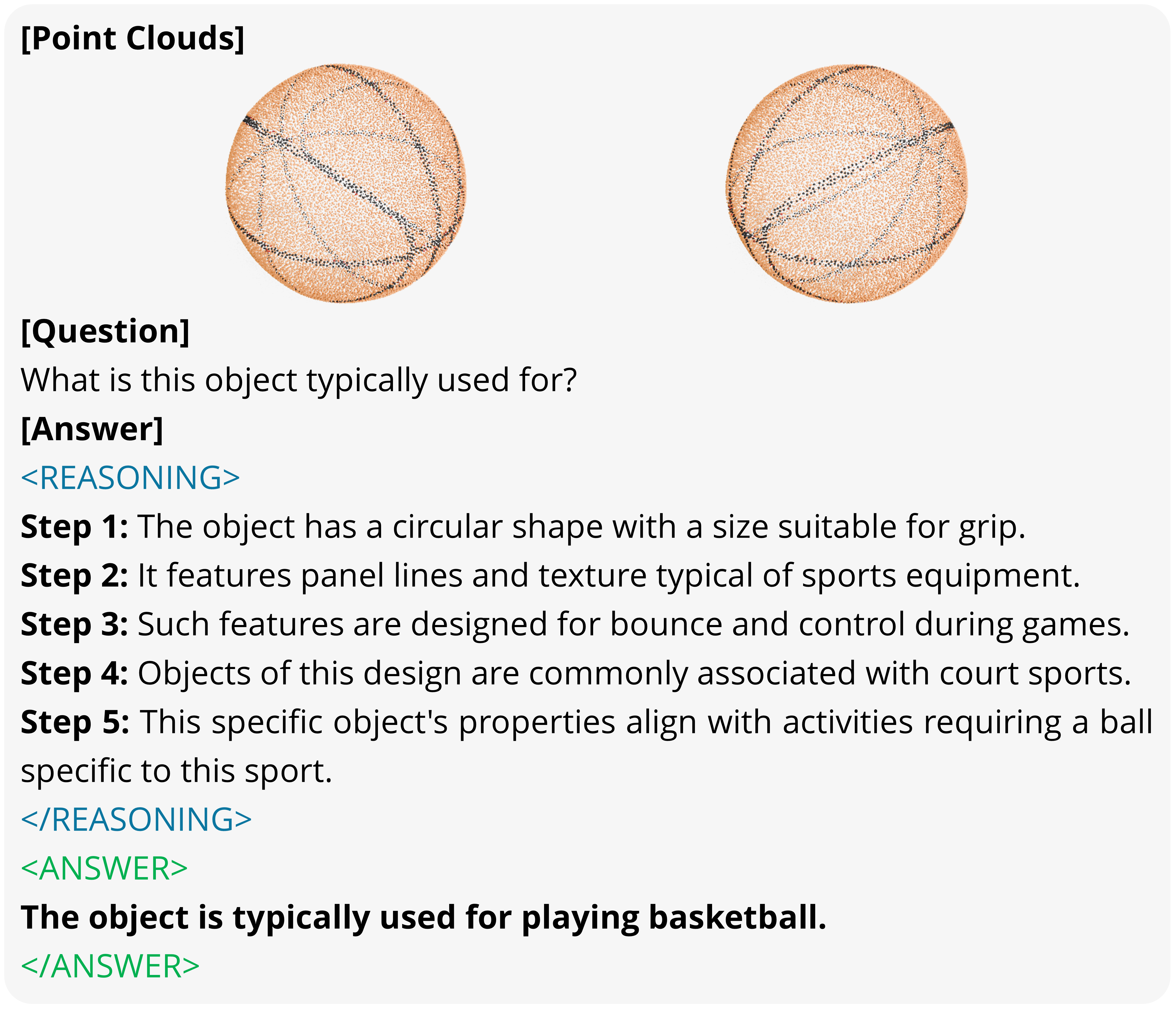}
    \vspace{-3mm}
    \caption{An illustrative example of our PoCoTI dataset, consisting of an input point cloud, a user query, the model's CoT reasoning, and the final answer.}
    \label{fig:data_sample}
    \end{center}
    \vspace{-5mm}
\end{figure}

\subsection{PoCoTI Data Generation}
\label{sec:HiLPO}

\subsubsection{Human-in-the-Loop Prompt Optimization}
Given $\mathcal{D}_{\text{refined}}$, we aim to construct a CoT generation prompt to bridge each $(I, A)$ pair with reasoning paths. However, designing an effective prompt for 3D CoT generation is non-trivial, as a manually crafted prompt often induces suboptimal or hallucinated reasoning. We observe that such failure cases expose systematic weaknesses in the guiding prompt, motivating a Human-in-the-Loop Prompt Optimization (HiLPO) framework that iteratively refines the prompt by combining LLM-driven feedback with expert oversight. The process is shown in Fig. ~\ref{fig:overview}, and comprises several core steps:

\paragraph{Prompt Initialization.}
We start from a manually crafted prompt $\mathcal{P}_0$ designed as a structured template with explicit reasoning constraints and strict output formatting, and set $\mathcal{P}_{\text{current}} \leftarrow \mathcal{P}_0$.

\paragraph{Data Sample Generation.}
Guided by $\mathcal{P}_{\text{current}}$, we employ Qwen3-VL as the vision-language model $L_V$ to generate CoT snippets by jointly processing $(V_P, I, A)$:
\begin{equation}
s_{k,j} = L_V(V_{P_j}, I_j, A_j, \mathcal{P}_{\text{current}}), \quad j=1,\dots,N_S,
\end{equation}
where $k$ denotes the current optimization iteration and $N_S = 100$ ensures representative prompt evaluation.

\paragraph{LLM-based Refinement and Human Verification.}
The generated samples $S_k$ and current prompt are analyzed by a refinement Claude\footnote{In this paper, Claude refers to Anthropic's ``claude-sonnet-4-20250514''} model $L_R$ to produce a candidate prompt, which is then selectively accepted or rejected by human experts to ensure task alignment and structural consistency:
\begin{equation}
\mathcal{P}_{\text{current}} \leftarrow \mathcal{H}(L_R(S_k, \mathcal{P}_{\text{current}})).
\end{equation}

The iterative optimization of the data generation prompt via HiLPO is visualized in Fig.~\ref{fig:CoT_Prompt}. The process converged after two successive iterations, with the refinements from the first and second rounds highlighted in \textcolor{myImageBlue}{blue} and \textcolor{myImageOrange}{orange}, respectively. Following these refinements, the prompt exhibits improved logical consistency and task-specific sensitivity, and is therefore adopted as the final prompt $P^*$ for large-scale dataset synthesis (see supplementary material for more details).

\rev{Different from standard prompt engineering, which often involves manual inspection of outputs and iterative prompt refinement, our HiLPO framework offloads large-scale sample analysis and candidate prompt generation to an LLM, while reserving human effort for expert verification and selection. 
This division of labor is particularly beneficial in settings where prompt quality needs to be evaluated over large datasets, enabling scalable and consistent assessment. 
By combining automated feedback aggregation with domain-expert validation, HiLPO reduces the manual effort required for prompt tuning, while preserving task-specific alignment.}

\subsubsection{Synthesis of the PoCoTI Dataset}
Utilizing the prompt $\mathcal{P}^*$, we scale the generation process and propose the \textit{CoT-Enhanced Point-Text Instruction Following} (PoCoTI) dataset, effectively mitigating the scarcity of high-quality 3D CoT data. 

Specifically, for each sample, the multi-view renderings $V$, instruction $I$, and ground-truth answer $A$ are concatenated with the optimized prompt $P^*$ and fed into the vision language model $L_V$:
\begin{equation}
    (R, A) = L_V(V, I, A, \mathcal{P}^*),
    \label{eq:cot_generation}
\end{equation}
where $R$ is the generated reasoning path.
Notably, the ground-truth $A$ in the input guides $L_V$ to derive a coherent path $R$ that justifies the known conclusion, ensuring that the synthesized CoT reasoning is both factually grounded and geometrically consistent.
Ultimately, this process yields the final PoCoTI dataset $D_{CoT}$, comprising approximately 55K $(P, I, R, A)$ samples, with a representative example shown in Fig.~\ref{fig:data_sample}. This comprehensive corpus serves as the primary training source for our model.

\subsection{PointLLM-R}
\label{sec:PointLLM-R}

We propose \textit{PointLLM-R}, a 3D multi-modal large language model with CoT reasoning ability over point cloud inputs, obtained by fine-tuning PointLLM on our PoCoTI dataset.
PointLLM-R adopts Point-BERT~\cite{yu2022point} (pretrained with ULIP-2~\cite{xue2024ulip} on Objaverse) as the point cloud encoder, whose parameters are frozen during training.
Only the projector and the language model are optimized, with trainable parameters denoted as $\theta = (\theta_{\text{proj}}, \theta_{\text{LLM}})$.

Given a sample $(P, I, R, A) \sim D_{\text{CoT}}$,
The optimization objective is to minimize the negative log-likelihood of the target output sequence $Y=R+A$.
The loss function $\mathcal{L}(\theta)$ is then computed as:
\begin{equation}
    \mathcal{L}(\theta) = - \sum_{(P, I, R, A) \sim D_{\text{CoT}}} \sum_{k=1}^{|Y|} \log P_{\theta}(\mathit{token}_k | \mathit{prefix}_k, P, I; \theta),
    \label{eq:training_loss}
\end{equation}
where $\mathit{token}_k$ is the $k$-th token in $Y$, and $\mathit{prefix}_k$ is the sequence of $(k-1)$ preceding tokens. 
Consequently, this fine-tuning strategy trains the model to first output reasoning steps before deriving the final answer, thereby enhancing its reasoning capability.


\section{Experiments}
\label{sec:exp}

In this section, we conduct empirical evaluations of the proposed PointLLM-R-7B model across two distinct tasks: \textit{Generative 3D point cloud classification} and \textit{3D point cloud captioning}.

\begin{table*}[t]
    \vspace{-3mm}
    \centering
    \caption{Generative 3D object classification results on ModelNet40 (M40.), Objaverse (Obj.), and OmniObject3D (Omni.) under a zero-shot setting.
    Results are evaluated using automatic LLM judging following the PointLLM protocol, with two prompt types: an instruction-style prompt (I, ``What is this?'') and a completion-style prompt (C, ``This is an object of'').
    Each entry reports accuracy averaged across multiple LLM judges, and the final column shows the average accuracy across all datasets and prompt types.
    Our model, PointLLM-R-7B, achieves state-of-the-art performance across all benchmarks.}

    \begin{tabular}{lccccccc}
        \toprule
        Model & M40.(I) & M40.(C) & Obj.(I) & Obj.(C) & Omni.(I) & Omni.(C) & Average \\
        \midrule    
        ShapeLLM-7B~\citep{qi2024shapellm} & 19.80 & 11.84 & 23.44 & 20.56 & 19.71 & 20.01 & 19.23 \\
        ShapeLLM-13B~\citep{qi2024shapellm}  & 22.17 & 20.15 & 32.28 & 33.47 & 19.95 & 20.31 & 24.72 \\
        PointLLM-7B~\citep{xu2024pointllm}  & 53.24 & 53.42 & 48.85 & 48.40 & 23.46 & 23.35 & 41.79 \\
        PointLLM-13B~\citep{xu2024pointllm}  & 53.79 & 53.99 & 49.83 & 49.78 & 23.82 & 23.83 & 42.51 \\
        MiniGPT-3D~\citep{tang2024minigpt}  & 57.49 & 57.63 & 52.90 & 52.63 & 25.99 & 26.14 & 45.46 \\
        \textbf{PointLLM-R-7B (Ours)} & \textbf{62.40} & \textbf{61.60} & \textbf{59.17} & \textbf{59.27} & \textbf{33.22} & \textbf{33.27} & \textbf{51.49} \\
        \midrule
        \end{tabular}
    \label{tab:classification}
\end{table*}
\begin{table*}[t]
    \vspace{-3mm}
    \centering
    
    \caption{3D object captioning results on the Objaverse dataset.
    Captions are evaluated using automatic LLM judging following the PointLLM evaluation protocol, with four LLM judges treated equally.
    The reported LLM-based scores include the individual judge scores and their average, and we additionally report text-based similarity metrics including Sentence-BERT and SimCSE.
    PointLLM-R-7B achieves the best performance across all metrics.}

    \begin{tabular}{lccccc|cc}
        \toprule
        \multirow{2}*{Model} & \multicolumn{5}{c|}{LLM-Based} & \multicolumn{2}{c}{Text-Based} \\
        \cline{2-8}
          & GPT-4 & Gemini-3 & Qwen-3 & GLM-4.6 & Average & Sentence-BERT & SimCSE \\
        \midrule
        ShapeLLM-7B~\citep{qi2024shapellm} & 22.35 & 24.71 & 23.56 & 19.84 & 22.53 & 37.82 & 38.84  \\
        ShapeLLM-13B~\citep{qi2024shapellm} & 29.64 & 31.05 & 30.84 & 26.94 & 29.62 & 39.72 & 41.56 \\
        PointLLM-7B~\citep{xu2024pointllm} & 46.71 & 54.31 & 49.40 & 41.84 & 48.07 & 46.99 & 47.61  \\
        PointLLM-13B~\citep{xu2024pointllm} & 48.05 & 55.34 & 51.09 & 43.89 & 49.59 & 47.65 & 48.35 \\
        MiniGPT-3D~\citep{tang2024minigpt} & 52.23 & 56.93 & 56.84 & 50.56 & 54.14 & 48.74 & 50.76  \\
        \textbf{PointLLM-R-7B (Ours)}  & \textbf{56.78} & \textbf{59.61} & \textbf{62.17} & \textbf{54.56} & \textbf{58.28} & \textbf{49.03} & \textbf{51.25} \\
        \bottomrule
    \end{tabular}
    \label{tab:captioning}
\end{table*}
\subsection{Experimental Setup}
\label{exp:setup}

\paragraph{Datasets.} 
Following the evaluation protocol of PointLLM~\cite{xu2024pointllm}, 
we utilize ModelNet-40~\cite{wu20153d}, Objaverse~\cite{deitke2023objaverse3}, and Cap3D~\cite{luo2023scalable}.
ModelNet-40 contains 12,311 synthetic 3D CAD models across 40 categories, and we use its test split of 2,468 point clouds for generative 3D object classification.
Objaverse offers over 800K 3D assets.
Cap3D provides more than 1M 3D-text pairs.
We use 3,000 Objaverse point clouds with their corresponding Cap3D captions as ground truth for both generative 3D object classification and 3D object captioning.

To incorporate real-world scanned data into the evaluation,
we include OmniObject3D~\cite{wu2023omniobject3d}, a real-scanned dataset covering 190 daily-use categories.
We extend the generative 3D object classification task with 5,898 objects from OmniObject3D.
As OmniObject3D lacks textual annotations, we render each object into four views and prompt GPT-5.2 to generate brief caption as ground truth.
All evaluation data are unseen during training.

\paragraph{Metrics.}
For classification, we adopt a generative evaluation protocol with two prompts: an instruction-style prompt (\textbf{I}: ``What is this?'') and a completion-style prompt (\textbf{C}: ``This is an object of''), and compute accuracy via automatic LLM judging. 
All evaluations use the prompt templates from PointLLM and are conducted with multiple LLM judges from different vendors (GPT-4, Qwen-3, Gemini-3, and GLM-4.6)\footnote{In this paper, GPT-4 refers to OpenAI's ``gpt-4-0613'', Qwen-3 refers to Alibaba's ``qwen-flash-2025-07-28''; Gemini-3 refers to Google's ``gemini-3-flash-preview''; and GLM-4.6 refers to Zhipu AI's ``glm-4.6''.}, which are treated equally and averaged to mitigate evaluator bias.
For ModelNet40, we perform strict zero-shot evaluation, where the judge selects exactly one label from the 40 categories based on the model response and the prediction is correct if it matches the ground truth.
For Objaverse and OmniObject3D, we adopt open-ended category matching, where the judge determines whether the response refers to the same category as the ground truth (binary T/F); OmniObject3D is also evaluated in a zero-shot setting to assess generalization to real-scanned objects.

For captioning, we prompt the model with ``Caption this 3D model in detail.'' 
Captions are scored by the same set of LLM judges using PointLLM's caption evaluation prompt.
We report the average LLM score in the main paper, while the individual scores from each judge are also provided for reference (in the main paper for captioning and in the supplementary material for classification).
We additionally report embedding-based similarity (Sentence-BERT~\cite{reimers2019sentence} and SimCSE~\cite{gao2021simcse} cosine similarity).

\paragraph{Baselines.}
We compare PointLLM-7B-R with several strong baselines capable of performing the same generative classification and captioning tasks. 
The primary baselines include PointLLM, MiniGPT-3D~\cite{tang2024minigpt}, and ShapeLLM~\cite{qi2024shapellm}, all of which are designed for multimodal 3D understanding with point cloud inputs. 
For ShapeLLM and PointLLM, we conduct experiments using both their 7B and 13B checkpoints.
These baselines serve as relevant comparisons for evaluating the effectiveness of our model.

\subsection{Generative 3D Classification}
\label{exp:results}

\paragraph{Quantitative Results} As reported in Tab.~\ref{tab:classification}, across all datasets and prompt types, PointLLM-R reaches state-of-the-art performance on generative 3D object classification, with an average accuracy of \textbf{51.49\%}.
PointLLM-R consistently outperforms prior 3D MLLMs across all benchmarks.
Our model achieves substantial gains under both instruction-style and completion-style prompts on the ModelNet40 dataset, indicating strong generalization to unseen categories.
Compared to PointLLM-7B, PointLLM-R improves the average accuracy by \textbf{+9.70\%}, demonstrating the effectiveness of CoT-enriched supervision generated by our framework.

Notably, the performance improvement is most pronounced on OmniObject3D, which consists of real-world scanned objects and poses greater challenges due to geometric noise, partial observations, and domain shift.
PointLLM-R surpasses all baselines by a large margin on both prompt types, highlighting its enhanced robustness and generalization ability when transferring from synthetic training data to real-scanned 3D objects.
These results suggest that reasoning-oriented data generation not only benefits synthetic benchmarks but also significantly improves real-world applicability.

We further observe that PointLLM-R outperforms larger 13B models, exceeding PointLLM-13B and ShapeLLM-13B by a clear margin on average.
This demonstrates that high-quality, CoT-enhanced instruction data can enable more compact models to achieve stronger 3D understanding and reasoning capability than larger counterparts trained without such supervision.

\begin{figure*}[t]
    \begin{center}
    \centering
    \includegraphics[width=0.95\textwidth]{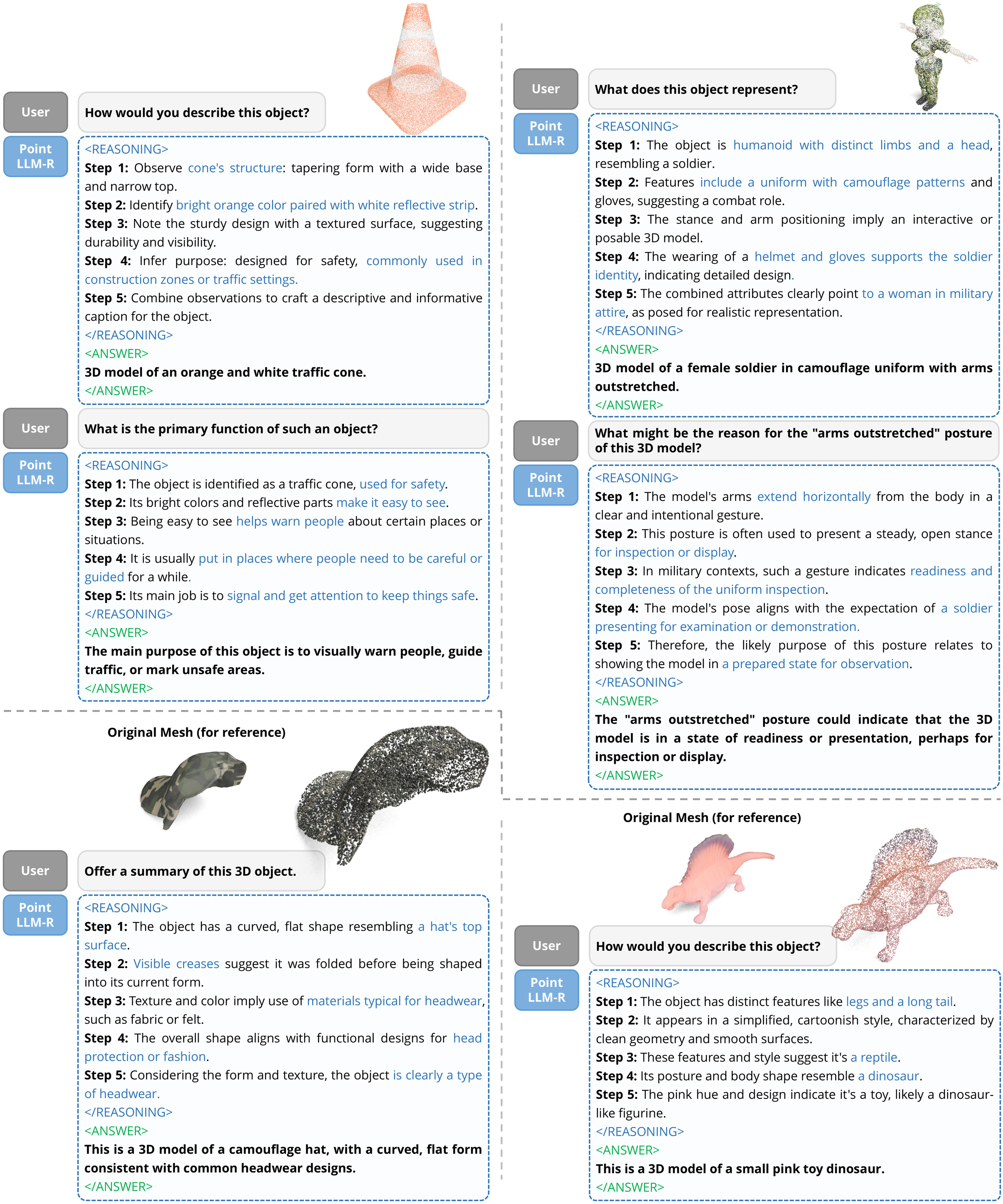}
    
    \caption{Examples of interactions with PointLLM-R. The top two panels demonstrate multi-round dialogues, where PointLLM-R addresses consecutive user queries for a given input point cloud. The bottom two panels present responses to single-turn user queries for real scanned point clouds from OmniObject3D dataset. The rendered mesh images are solely for visual reference here and do not constitute input data.}
    \label{fig:multi_turn_and_real_scan}
    \end{center}
\end{figure*}

\paragraph{Qualitative Results}
We present qualitative examples on objects from real-scanned OmniObject3D dataset to illustrate the behavior of PointLLM-R in challenging real-world scenarios.
As shown in Fig.~\ref{fig:multi_turn_and_real_scan}, despite geometric noise, incomplete surfaces, and non-uniform point density, PointLLM-R is able to infer semantically meaningful object categories and produce coherent descriptions grounded in the underlying 3D structure.

In the first example, the input point cloud exhibits irregular geometry and partial observations, yet the model successfully captures fine-grained geometric cues such as surface curvature and visible crease patterns, which support a coherent interpretation of the object’s shape and functional category.
In the second example, although the object geometry is highly simplified and stylized, PointLLM-R successfully reasons about distinctive structural features such as body shape and appendages, and identifies the object as a toy-like instance with appropriate semantic attributes.

These examples demonstrate that PointLLM-R can effectively leverage 3D geometric cues to support structured reasoning and semantic inference on real-scanned data, even in the presence of significant noise and domain shift.
This further validates the robustness and practical applicability of our CoT-enhanced 3D reasoning framework beyond synthetic benchmarks.

\subsection{3D Object Captioning}

\paragraph{Quantitative Results.}
Tab.~\ref{tab:captioning} reports the quantitative results of 3D object captioning on Objaverse.
Although different LLM judges exhibit varying scoring preferences and inherent biases, PointLLM-R consistently achieves the best performance across all individual judges.
This consistent superiority across evaluators indicates that the performance gains of PointLLM-R are not tied to any specific judge, but instead reflect robust and broadly aligned improvements in caption quality.

In terms of the averaged LLM-based score, PointLLM-R surpasses all prior methods by a clear margin, improving over the strongest baseline MiniGPT-3D by \textbf{+4.14} points.
Beyond LLM-based evaluation, PointLLM-R also achieves the highest scores on text-based similarity metrics, including Sentence-BERT and SimCSE.

Compared to PointLLM and ShapeLLM with larger parameter sizes, PointLLM-R consistently delivers stronger results across all reported metrics.
These results demonstrate that reasoning-oriented supervision enables a compact model to produce accurate, detailed, and semantically grounded 3D captions, yielding stable performance gains across diverse evaluators.

\begin{figure*}[t]
    \begin{center}
    \centering
    \includegraphics[width=0.95\textwidth]{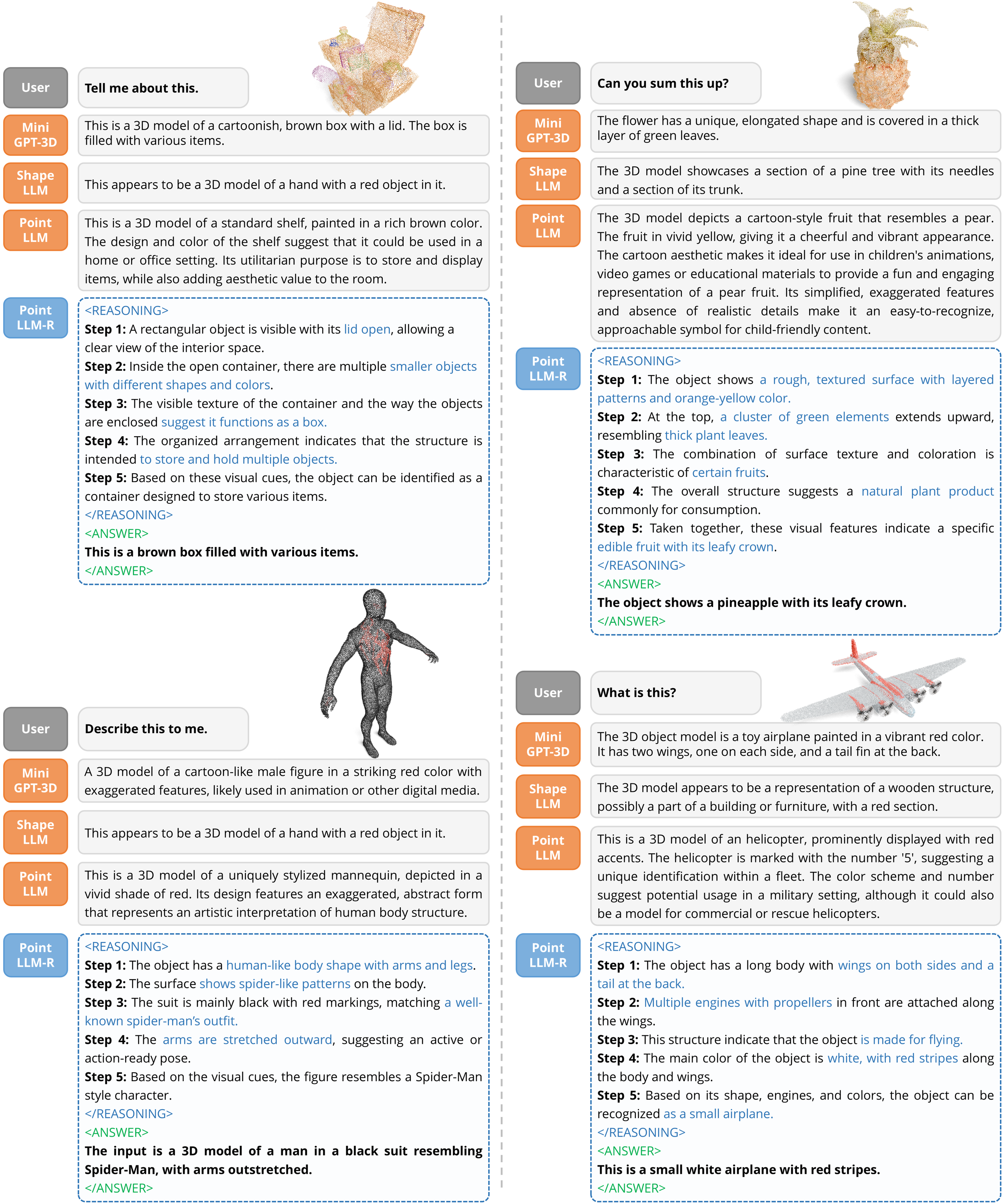}
    \caption{Qualitative examples illustrating responses from various models to user prompts for four distinct 3D object. Each example begins with the user's textual query and a point cloud input. This is followed by the respective textual outputs from MiniGPT-3D, ShapeLLM, and PointLLM. Finally, the detailed CoT reasoning process and the resulting textual answer from PointLLM-R are displayed.}
    \label{fig:qualitative_result}
    \end{center}
\end{figure*}

\paragraph{Qualitative Results.}
Fig.~\ref{fig:qualitative_result} presents qualitative comparisons between our model and baseline models
Across all cases, PointLLM-R generates more accurate and structured descriptions by explicitly reasoning over fine-grained geometric cues.

A particularly illustrative example is shown in the pineapple case.
While baseline models fail to correctly recognize the object, describing it as a generic fruit, a pear-like shape, or even a tree-related structure, PointLLM-R successfully identifies it as a pineapple.
This is achieved by decomposing the point cloud into interpretable cues, such as the rough, layered surface, the orange-yellow coloration, and the characteristic crown of leaves, and integrating them through step-by-step reasoning.

Similar patterns can be observed in the other examples, where PointLLM-R analyzes object parts, spatial arrangement, and functional attributes to produce concise yet semantically grounded captions.
In contrast, baseline methods tend to generate brief or ambiguous descriptions that overlook critical geometric details.
These qualitative results demonstrate that CoT supervision enables more precise object recognition and more interpretable 3D understanding from raw point clouds.

\begin{table}[!tbp]
    \centering
    \caption{\rev{Ablation study on data sources for constructing the CoT training data,
    analyzing the contribution of ShapeLLM SFT data and Cap3D captions to final model performance.}}
    \vspace{-3mm}
    \begin{tabular}{cccc}
        \toprule
        w. Shape. & w. Cap3D  & Cls. Acc. & Cap. Acc. \\
        \midrule
        \ding{55} & \ding{55} & 41.79 & 48.07 \\
        \ding{55} & \ding{51} & 43.98 & 50.31 \\
        \ding{51} & \ding{55} & 50.88 & 57.87 \\
        \ding{51} & \ding{51} & \textbf{51.49} & \textbf{58.28} \\
        \bottomrule
    \vspace{-5mm}
    \end{tabular}
    \label{tab:data_source_ablation}
\end{table}
\begin{table}[!tbp]
    \centering
    \caption{\rev{Ablation study on different stages of the data generation pipeline,
    analyzing data refinement and Human-in-the-Loop Prompt Optimization (HiLPO) under a fixed CoT generation setting.}}
    \begin{tabular}{ccccc}
        \toprule
        w. Refinement  & w. HiLPO & w. CoT & Cls. Acc. & Cap. Acc. \\
        \midrule
        \ding{55} & \ding{55} & \ding{51} & 41.30 & 49.63 \\
        \ding{51} & \ding{55} & \ding{55} & 44.42 & 48.38 \\
        \ding{55} & \ding{51} & \ding{51} & 49.69 & 55.84 \\
        \ding{51} & \ding{55} & \ding{51} & 46.32 & 54.32 \\
        \ding{51} & \ding{51} & \ding{51} & \textbf{51.49} & \textbf{58.28} \\
        \bottomrule
    \vspace{-5mm}
    \end{tabular}  
    \label{tab:data_gen_stage_ablation}
\end{table}

\subsection{Ablation Study}
\paragraph{Effect of Data Sources for CoT Construction.}
We first investigate how different data sources contribute to the quality of the constructed CoT training data.
\rev{As shown in Table~\ref{tab:data_source_ablation}, when neither ShapeLLM SFT data nor Cap3D captions are used, the model achieves 41.79 classification accuracy and 48.07 captioning accuracy.
Using Cap3D captions alone increases the scores to 43.98 and 50.31, corresponding to gains of 2.19 and 2.24 points, respectively.
Using ShapeLLM SFT data alone leads to substantially larger improvements, reaching 50.88 classification accuracy and 57.87 captioning accuracy, which are 9.09 and 9.80 points higher than the no-data-source setting.
Combining both data sources yields the best performance, further improving the results to \textbf{51.49} classification accuracy and \textbf{58.28} captioning accuracy.
These results show that ShapeLLM SFT data provides the major performance gains, while Cap3D captions offer additional complementary benefits, and together they help construct higher-quality CoT training data for robust 3D point cloud reasoning.}

\rev{\paragraph{Effect of Data Generation Pipeline Stages.}
We further analyze the contributions of different stages in our data generation pipeline.
As reported in Table~\ref{tab:data_gen_stage_ablation}, the full pipeline achieves the best performance, reaching \textbf{51.49} classification accuracy and \textbf{58.28} captioning accuracy.
Removing data refinement while keeping HiLPO reduces the performance to 49.69 classification accuracy and 55.84 captioning accuracy, corresponding to drops of 1.80 and 2.44 points, respectively.
Removing HiLPO while keeping data refinement leads to a larger decline, with classification accuracy decreasing to 46.32 and captioning accuracy to 54.32, i.e., drops of 5.17 and 3.96 points, respectively.
In particular, the larger performance gap caused by disabling HiLPO indicates that prompt quality is a key factor in determining the usefulness of the synthesized CoT supervision.
The worst results are obtained when both data refinement and HiLPO are removed, further confirming that these components are complementary and jointly critical for constructing reliable CoT supervision.}

\begin{figure}[!tbp]
    \begin{center}
    \centering
    \vspace{-3mm}
    \includegraphics[width=0.42\textwidth]{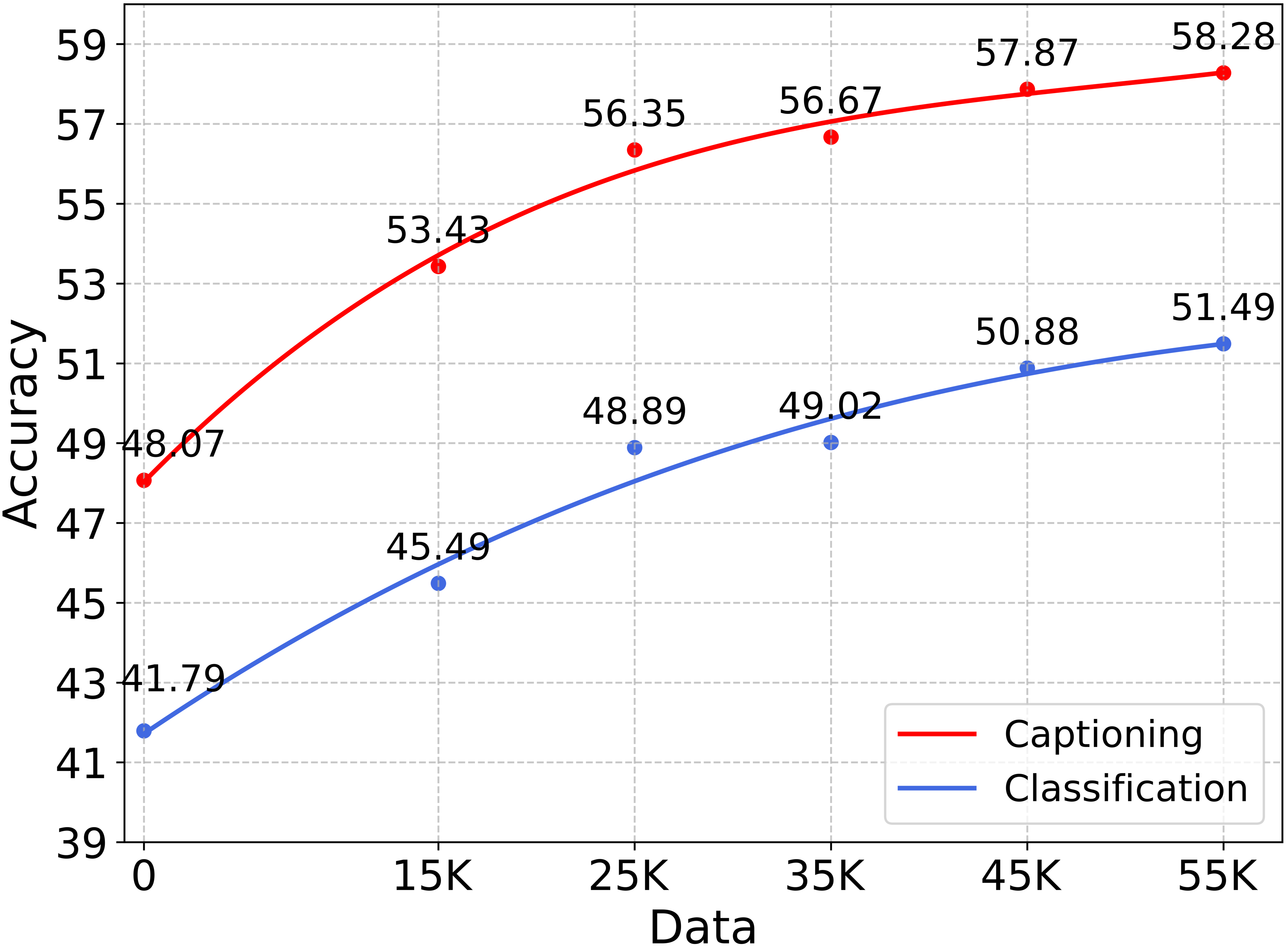}
    \caption{Accuracy of PointLLM-R with varying sizes of the PoCoTI dataset during fine-tuning process in classification and caption tasks, where the X-axis denotes the number of training samples.}
    \label{fig:data_ablation}
    \vspace{-5mm}
    \end{center}
\end{figure}

\paragraph{Effect of CoT Dataset Scale.}
Finally, we study the impact of the size of the constructed PoCoTI dataset.
Figure~\ref{fig:data_ablation} illustrates the performance of PointLLM-R as the number of training samples increases.
We observe a clear and consistent performance gain on both classification and captioning tasks as more CoT training data are used.
This trend confirms that PointLLM-R effectively leverages large-scale, CoT-enriched instruction data, and further validates the scalability and effectiveness of our data generation pipeline.

\section{Conclusion}

This paper addresses the challenge of enabling CoT reasoning in 3D multimodal large language models under limited high-quality reasoning supervision.
We propose a data-centric solution that systematically constructs CoT supervision through a two-stage pipeline, consisting of data refinement and HiLPO.
Based on this pipeline, we build PoCoTI, a large-scale CoT-enhanced point-text instruction dataset with explicit, geometrically grounded reasoning paths.

Fine-tuning PointLLM on PoCoTI, we present PointLLM-R, a 3D MLLM with strong CoT reasoning capability over point cloud inputs.
Extensive experiments and ablation studies demonstrate that our data construction strategy effectively improves reasoning performance and generalization across generative 3D classification and captioning tasks, validating the scalability and effectiveness of the proposed approach.

\begin{figure}[t]
    \begin{center}
    \centering
    \includegraphics[width=0.47\textwidth]{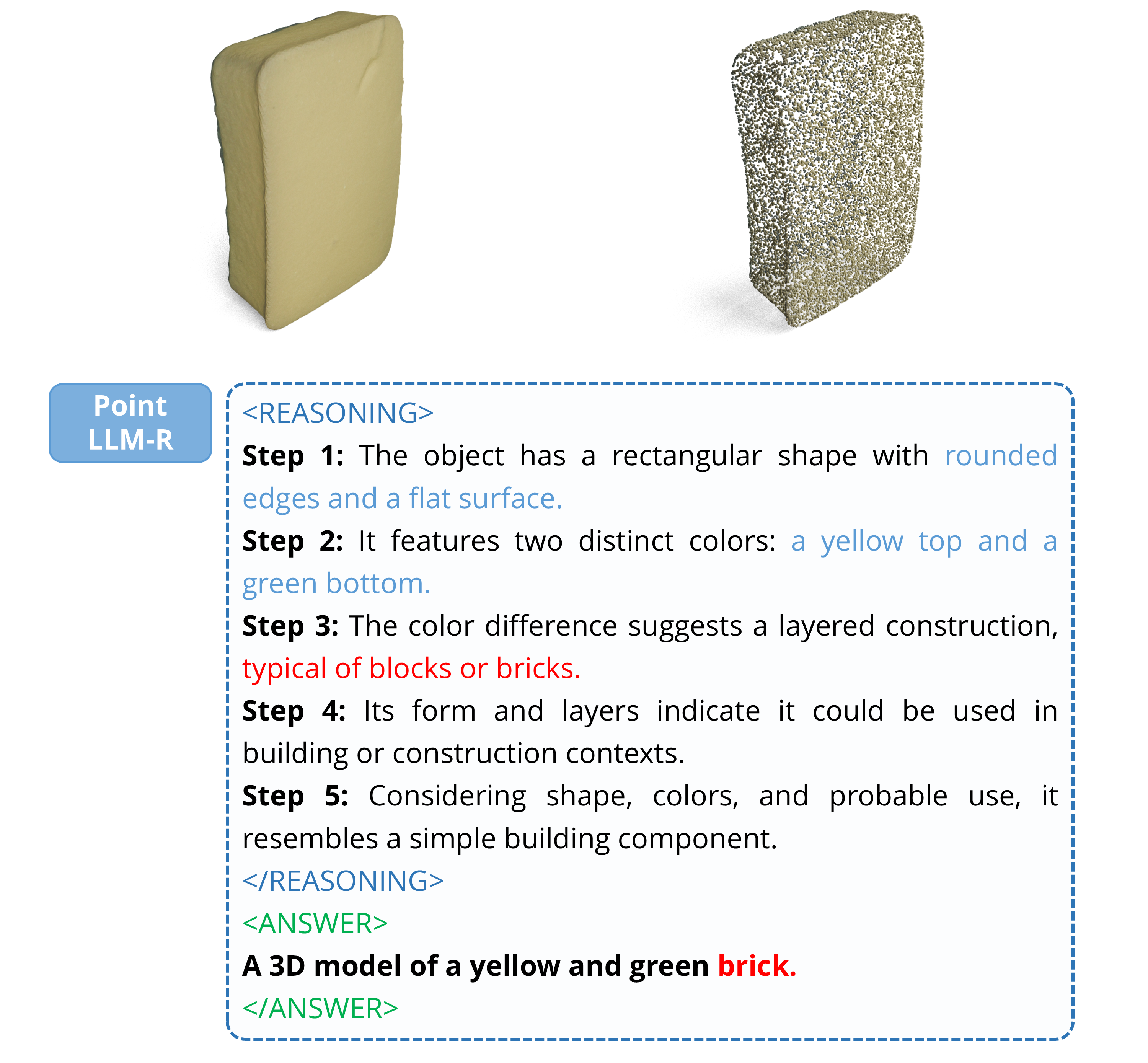}
    \caption{
    \rev{Failure case of our method. The model captures correct shape and color cues but misclassifies the sponge as a brick due to insufficient fine-grained evidence. 
    The reasoning remains internally consistent but leads to a plausible yet incorrect conclusion.}
    }
    \label{fig:failure_case}
    \end{center}
\end{figure}

\rev{\paragraph{Limitations}
As shown in Fig.~\ref{fig:failure_case}, our model may produce incorrect predictions when the input point cloud provides insufficient discriminative evidence due to sparse sampling. 
In such cases, objects with similar global appearance but different semantic categories can be confused, as the available observations do not capture fine-grained cues necessary for disambiguation. 
Importantly, the reasoning process itself remains logically consistent, suggesting that the error primarily stems from limitations in the input signal rather than the reasoning mechanism.

In future work, we plan to further improve the model’s ability to handle ambiguous cases by enriching the training data with more challenging samples that exhibit subtle distinctions between similar object categories. We will also explore extending our CoT dataset to cover more complex reasoning scenarios, diversified multimodal contexts, and more challenging task structures. Additionally, we will investigate adapting our model to handle complex, scene-level 3D point clouds, further enhancing its generalization and robustness in real-world applications.}


\begin{acks}
This work was supported in part by Guangdong Science and Technology Program (2024B0101050004), ICFCRT (W2441020), Guangdong Basic and Applied Basic Research Foundation (2023B1515120026), Shenzhen Science and Technology Program (KJZD20240903100022028, KQTD20210811090044003), and Scientific Development Funds from Shenzhen University.
\end{acks}

\bibliographystyle{ACM-Reference-Format}
\bibliography{ref}

@String(CVPR= {IEEE Conf. Comput. Vis. Pattern Recog.})

@String(ICCV= {Int. Conf. Comput. Vis.})

@String(ECCV= {Eur. Conf. Comput. Vis.})

@String(EMNLP= {Conf. Empir. Methods Nat. Lang. Process.})

@String(EMNLP_IJCNLP  = {Conf. Empir. Methods Nat. Lang. Process. & Int. Joint Conf. Nat. Lang. Process.})

@String(NIPS= {Adv. Neural Inform. Process. Syst.})

@String(TOG= {ACM Trans. Graph.})

@String(TVCG  = {IEEE Trans. Vis. Comput. Graph.})

@String(ACMMM= {ACM Int. Conf. Multimedia})

@String(ICLR = {Int. Conf. Learn. Represent.})

@String(ICRA = {IEEE Int. Conf. Robot. Autom.})

@String(AAAI = {AAAI})

@String(COLING = {Int. Conf. Comput. Linguist.})

@String(NAACL= {Conf. North Am. Chapter Assoc. Comput. Linguistics})

@String(CVPR  = {CVPR})

@String(ICCV  = {ICCV})

@String(ECCV  = {ECCV})

@String(EMNLP  = {EMNLP})

@String(EMNLP_IJCNLP  = {EMNLP\_IJCNLP})

@String(COLING  = {COLING})

@String(NIPS  = {NeurIPS})

@String(NAACL  = {NAACL})

@String(TOG   = {ACM TOG})

@String(TVCG  = {IEEE TVCG})

@String(ACMMM = {ACM MM})

@String(ICLR  = {ICLR})

@String(ICRA = {ICRA})

@String(IROS = {IROS})

@STRING(TOG = {ACM Trans. on Graphics})

@STRING(TOG_SIG = {ACM Trans. on Graphics (Proc. SIGGRAPH)})

@STRING(ICML = {Proc. Int. Conf. on Machine Learning})

@String(TMLR = {Trans. Mach. Learn. Res.})

@article{wang2025chat2layout,
  title={Chat2Layout: Interactive 3D furniture layout with a multimodal LLM},
  author={Wang, Can and Zhong, Hongliang and Chai, Menglei and He, Mingming and Chen, Dongdong and Liao, Jing},
  journal=TVCG,
  year={2025},
  publisher={IEEE}
}

@article{tang2025lego,
  title={Lego-puzzles: How good are mllms at multi-step spatial reasoning?},
  author={Tang, Kexian and Gao, Junyao and Zeng, Yanhong and Duan, Haodong and Sun, Yanan and Xing, Zhening and Liu, Wenran and Lyu, Kaifeng and Chen, Kai},
  journal={CoRR},
  year={2025}
}

@article{yuan2025scene,
  title={Scene-r1: Video-grounded large language models for 3d scene reasoning without 3d annotations},
  author={Yuan, Zhihao and Jiang, Shuyi and Feng, Chun-Mei and Zhang, Yaolun and Cui, Shuguang and Li, Zhen and Zhao, Na},
  journal={CoRR},
  year={2025}
}

@inproceedings{linghu2026scenecot,
  title={SceneCOT: Eliciting Grounded Chain-of-Thought Reasoning in 3D Scenes},
  author={Linghu, Xiongkun and Huang, Jiangyong and Zhu, Ziyu and Jia, Baoxiong and Huang, Siyuan},
  booktitle=ICLR,
  year={2026}
}

@inproceedings{maturana2015voxnet,
  title={Voxnet: A 3d convolutional neural network for real-time object recognition},
  author={Maturana, Daniel and Scherer, Sebastian},
  booktitle=IROS,
  pages={922--928},
  year={2015},
  organization={Ieee}
}

@inproceedings{qi2017pointnet,
  title={Pointnet: Deep learning on point sets for 3d classification and segmentation},
  author={Qi, Charles R and Su, Hao and Mo, Kaichun and Guibas, Leonidas J},
  booktitle=CVPR,
  pages={652--660},
  year={2017}
}

@article{qi2017pointnet++,
  title={Pointnet++: Deep hierarchical feature learning on point sets in a metric space},
  author={Qi, Charles Ruizhongtai and Yi, Li and Su, Hao and Guibas, Leonidas J},
  journal=NIPS,
  volume={30},
  year={2017}
}

@inproceedings{wu2023omniobject3d,
  title={Omniobject3d: Large-vocabulary 3d object dataset for realistic perception, reconstruction and generation},
  author={Wu, Tong and Zhang, Jiarui and Fu, Xiao and Wang, Yuxin and Ren, Jiawei and Pan, Liang and Wu, Wayne and Yang, Lei and Wang, Jiaqi and Qian, Chen and others},
  booktitle=CVPR,
  pages={803--814},
  year={2023}
}

@article{bai2025qwen3vl,
      title={Qwen3-VL Technical Report}, 
      author={Shuai Bai and Yuxuan Cai and Ruizhe Chen and Keqin Chen and Xionghui Chen and Zesen Cheng and Lianghao Deng and Wei Ding and Chang Gao and Chunjiang Ge and Wenbin Ge and Zhifang Guo and Qidong Huang and Jie Huang and Fei Huang and Binyuan Hui and Shutong Jiang and Zhaohai Li and Mingsheng Li and Mei Li and Kaixin Li and Zicheng Lin and Junyang Lin and Xuejing Liu and Jiawei Liu and Chenglong Liu and Yang Liu and Dayiheng Liu and Shixuan Liu and Dunjie Lu and Ruilin Luo and Chenxu Lv and Rui Men and Lingchen Meng and Xuancheng Ren and Xingzhang Ren and Sibo Song and Yuchong Sun and Jun Tang and Jianhong Tu and Jianqiang Wan and Peng Wang and Pengfei Wang and Qiuyue Wang and Yuxuan Wang and Tianbao Xie and Yiheng Xu and Haiyang Xu and Jin Xu and Zhibo Yang and Mingkun Yang and Jianxin Yang and An Yang and Bowen Yu and Fei Zhang and Hang Zhang and Xi Zhang and Bo Zheng and Humen Zhong and Jingren Zhou and Fan Zhou and Jing Zhou and Yuanzhi Zhu and Ke Zhu},
	   journal={CoRR},
      year={2025}
}

@inproceedings{tang2024minigpt,
      title={Minigpt-3d: Efficiently aligning 3d point clouds with large language models using 2d priors},
      author={Tang, Yuan and Han, Xu and Li, Xianzhi and Yu, Qiao and Hao, Yixue and Hu, Long and Chen, Min},
      booktitle=ACMMM,
      pages={6617--6626},
      year={2024}
}

@article{wang2017cnn,
  author       = {Peng{-}Shuai Wang and
                  Yang Liu and
                  Yu{-}Xiao Guo and
                  Chun{-}Yu Sun and
                  Xin Tong},
  title        = {{O-CNN:} octree-based convolutional neural networks for 3D shape analysis},
  journal      = TOG_SIG,
  volume       = {36},
  number       = {4},
  pages        = {72:1--72:11},
  year         = {2017},
}

@article{wang2019dynamic,
  author       = {Yue Wang and
                  Yongbin Sun and
                  Ziwei Liu and
                  Sanjay E. Sarma and
                  Michael M. Bronstein and
                  Justin M. Solomon},
  title        = {Dynamic Graph {CNN} for Learning on Point Clouds},
  journal      = TOG,
  volume       = {38},
  number       = {5},
  pages        = {146:1--146:12},
  year         = {2019},
}

@inproceedings{shen2024aligning,
  author       = {Yunhang Shen and
                  Chaoyou Fu and
                  Peixian Chen and
                  Mengdan Zhang and
                  Ke Li and
                  Xing Sun and
                  Yunsheng Wu and
                  Shaohui Lin and
                  Rongrong Ji},
  title        = {Aligning and Prompting Everything All at Once for Universal Visual
                  Perception},
  booktitle    = CVPR,
  pages        = {13193--13203},
  year         = {2024},
}

@inproceedings{yang2025storyllava,
  author       = {Li Yang and
                  Zhiding Xiao and
                  Wenxin Huang and
                  Xian Zhong},
  title        = {StoryLLaVA: Enhancing Visual Storytelling with Multi-Modal Large Language
                  Models},
  booktitle    = COLING,
  pages        = {3936--3951},
  year         = {2025},
}

@inproceedings{yan2025tg,
  author       = {Dawei Yan and
                  Pengcheng Li and
                  Yang Li and
                  Hao Chen and
                  Qingguo Chen and
                  Weihua Luo and
                  Wei Dong and
                  Qingsen Yan and
                  Haokui Zhang and
                  Chunhua Shen},
  title        = {TG-LLaVA: Text Guided LLaVA via Learnable Latent Embeddings},
  booktitle    = AAAI,
  pages        = {9076--9084},
  year         = {2025},
}

@inproceedings{li2025lrm,
  author       = {Junchen Li and
                  Qing Yang and
                  Bojian Jiang and
                  Shaolin Zhu and
                  Qingxuan Sun},
  title        = {LRM-LLaVA: Overcoming the Modality Gap of Multilingual Large Language-Vision
                  Model for Low-Resource Languages},
  booktitle    = AAAI,
  pages        = {24449--24457},
  year         = {2025},
}

@inproceedings{huang2024audiogpt,
  author       = {Rongjie Huang and
                  Mingze Li and
                  Dongchao Yang and
                  Jiatong Shi and
                  Xuankai Chang and
                  Zhenhui Ye and
                  Yuning Wu and
                  Zhiqing Hong and
                  Jiawei Huang and
                  Jinglin Liu and
                  Yi Ren and
                  Yuexian Zou and
                  Zhou Zhao and
                  Shinji Watanabe},
  title        = {AudioGPT: Understanding and Generating Speech, Music, Sound, and Talking
                  Head},
  booktitle    = AAAI,
  pages        = {23802--23804},
  year         = {2024},
}

@inproceedings{ma2025language,
  author       = {Ziyang Ma and
                  Yakun Song and
                  Chenpeng Du and
                  Jian Cong and
                  Zhuo Chen and
                  Yuping Wang and
                  Yuxuan Wang and
                  Xie Chen},
  title        = {Language Model Can Listen While Speaking},
  booktitle    = AAAI,
  pages        = {24831--24839},
  year         = {2025},
}

@inproceedings{tang2025empowering,
  author       = {Yunlong Tang and
                  Daiki Shimada and
                  Jing Bi and
                  Mingqian Feng and
                  Hang Hua and
                  Chenliang Xu},
  title        = {Empowering LLMs with Pseudo-Untrimmed Videos for Audio-Visual Temporal
                  Understanding},
  booktitle    = AAAI,
  pages        = {7293--7301},
  year         = {2025},
}

@inproceedings{chen2024videollm,
  author       = {Joya Chen and
                  Zhaoyang Lv and
                  Shiwei Wu and
                  Kevin Qinghong Lin and
                  Chenan Song and
                  Difei Gao and
                  Jia{-}Wei Liu and
                  Ziteng Gao and
                  Dongxing Mao and
                  Mike Zheng Shou},
  title        = {VideoLLM-online: Online Video Large Language Model for Streaming Video},
  booktitle    = CVPR,
  pages        = {18407--18418},
  year         = {2024},
}

@inproceedings{li2024llama,
  author       = {Yanwei Li and
                  Chengyao Wang and
                  Jiaya Jia},
  title        = {LLaMA-VID: An Image is Worth 2 Tokens in Large Language Models},
  booktitle    = ECCV,
  volume       = {15104},
  pages        = {323--340},
  year         = {2024},
}

@inproceedings{zhang2023video,
  title={Video-llama: An instruction-tuned audio-visual language model for video understanding},
  author={Zhang, Hang and Li, Xin and Bing, Lidong},
  booktitle={EMNLP},
  pages={543--553},
  year={2023}
}

@inproceedings{hong20233d,
  author       = {Yining Hong and
                  Haoyu Zhen and
                  Peihao Chen and
                  Shuhong Zheng and
                  Yilun Du and
                  Zhenfang Chen and
                  Chuang Gan},
  title        = {3D-LLM: Injecting the 3D World into Large Language Models},
  booktitle    = NIPS,
  year         = {2023},
}

@inproceedings{kirillov2023segment,
  author       = {Alexander Kirillov and
                  Eric Mintun and
                  Nikhila Ravi and
                  Hanzi Mao and
                  Chlo{\'{e}} Rolland and
                  Laura Gustafson and
                  Tete Xiao and
                  Spencer Whitehead and
                  Alexander C. Berg and
                  Wan{-}Yen Lo and
                  Piotr Doll{\'{a}}r and
                  Ross B. Girshick},
  title        = {Segment Anything},
  booktitle    = ICCV,
  pages        = {3992--4003},
  year         = {2023},
}

@inproceedings{zhang2022dino,
  author       = {Hao Zhang and
                  Feng Li and
                  Shilong Liu and
                  Lei Zhang and
                  Hang Su and
                  Jun Zhu and
                  Lionel M. Ni and
                  Heung{-}Yeung Shum},
  title        = {{DINO:} {DETR} with Improved DeNoising Anchor Boxes for End-to-End
                  Object Detection},
  booktitle    = ICLR,
  year         = {2023},
}

@article{oquab2023dinov2,
  author       = {Maxime Oquab and
                  Timoth{\'{e}}e Darcet and
                  Th{\'{e}}o Moutakanni and
                  Huy V. Vo and
                  Marc Szafraniec and
                  Vasil Khalidov and
                  Pierre Fernandez and
                  Daniel Haziza and
                  Francisco Massa and
                  Alaaeldin El{-}Nouby and
                  Mido Assran and
                  Nicolas Ballas and
                  Wojciech Galuba and
                  Russell Howes and
                  Po{-}Yao Huang and
                  Shang{-}Wen Li and
                  Ishan Misra and
                  Michael Rabbat and
                  Vasu Sharma and
                  Gabriel Synnaeve and
                  Hu Xu and
                  Herv{\'{e}} J{\'{e}}gou and
                  Julien Mairal and
                  Patrick Labatut and
                  Armand Joulin and
                  Piotr Bojanowski},
  title        = {DINOv2: Learning Robust Visual Features without Supervision},
  journal      = TMLR,
  volume       = {2024},
  year         = {2024},
}

@inproceedings{ji2024chain,
  author       = {Bin Ji and
                  Huijun Liu and
                  Mingzhe Du and
                  See{-}Kiong Ng},
  title        = {Chain-of-Thought Improves Text Generation with Citations in Large
                  Language Models},
  booktitle    = AAAI,
  pages        = {18345--18353},
  year         = {2024},
}

@inproceedings{wu20153d,
  author       = {Zhirong Wu and
                  Shuran Song and
                  Aditya Khosla and
                  Fisher Yu and
                  Linguang Zhang and
                  Xiaoou Tang and
                  Jianxiong Xiao},
  title        = {3D ShapeNets: {A} deep representation for volumetric shapes},
  booktitle    = CVPR,
  pages        = {1912--1920},
  year         = {2015},
}

@inproceedings{yu2022point,
  author       = {Xumin Yu and
                  Lulu Tang and
                  Yongming Rao and
                  Tiejun Huang and
                  Jie Zhou and
                  Jiwen Lu},
  title        = {Point-BERT: Pre-training 3D Point Cloud Transformers with Masked Point
                  Modeling},
  booktitle    = CVPR,
  pages        = {19291--19300},
  year         = {2022},
}

@article{guo2025deepseek,
  author       = {DeepSeek{-}AI and
                  Daya Guo and
                  Dejian Yang and
                  Haowei Zhang and
                  Junxiao Song and
                  Ruoyu Zhang and
                  Runxin Xu and
                  Qihao Zhu and
                  Shirong Ma and
                  Peiyi Wang and
                  Xiao Bi and
                  Xiaokang Zhang and
                  Xingkai Yu and
                  Yu Wu and
                  Z. F. Wu and
                  Zhibin Gou and
                  Zhihong Shao and
                  Zhuoshu Li and
                  Ziyi Gao and
                  Aixin Liu and
                  Bing Xue and
                  Bingxuan Wang and
                  Bochao Wu and
                  Bei Feng and
                  Chengda Lu and
                  Chenggang Zhao and
                  Chengqi Deng and
                  Chenyu Zhang and
                  Chong Ruan and
                  Damai Dai and
                  Deli Chen and
                  Dongjie Ji and
                  Erhang Li and
                  Fangyun Lin and
                  Fucong Dai and
                  Fuli Luo and
                  Guangbo Hao and
                  Guanting Chen and
                  Guowei Li and
                  H. Zhang and
                  Han Bao and
                  Hanwei Xu and
                  Haocheng Wang and
                  Honghui Ding and
                  Huajian Xin and
                  Huazuo Gao and
                  Hui Qu and
                  Hui Li and
                  Jianzhong Guo and
                  Jiashi Li and
                  Jiawei Wang and
                  Jingchang Chen and
                  Jingyang Yuan and
                  Junjie Qiu and
                  Junlong Li and
                  J. L. Cai and
                  Jiaqi Ni and
                  Jian Liang and
                  Jin Chen and
                  Kai Dong and
                  Kai Hu and
                  Kaige Gao and
                  Kang Guan and
                  Kexin Huang and
                  Kuai Yu and
                  Lean Wang and
                  Lecong Zhang and
                  Liang Zhao and
                  Litong Wang and
                  Liyue Zhang and
                  Lei Xu and
                  Leyi Xia and
                  Mingchuan Zhang and
                  Minghua Zhang and
                  Minghui Tang and
                  Meng Li and
                  Miaojun Wang and
                  Mingming Li and
                  Ning Tian and
                  Panpan Huang and
                  Peng Zhang and
                  Qiancheng Wang and
                  Qinyu Chen and
                  Qiushi Du and
                  Ruiqi Ge and
                  Ruisong Zhang and
                  Ruizhe Pan and
                  Runji Wang and
                  R. J. Chen and
                  R. L. Jin and
                  Ruyi Chen and
                  Shanghao Lu and
                  Shangyan Zhou and
                  Shanhuang Chen and
                  Shengfeng Ye and
                  Shiyu Wang and
                  Shuiping Yu and
                  Shunfeng Zhou and
                  Shuting Pan and
                  S. S. Li},
  title        = {DeepSeek-R1: Incentivizing Reasoning Capability in LLMs via Reinforcement
                  Learning},
  journal      = {CoRR},
  volume       = {abs/2501.12948},
  year         = {2025},
}

@article{bai2025qwen2.5vl,
  author       = {Shuai Bai and
                  Keqin Chen and
                  Xuejing Liu and
                  Jialin Wang and
                  Wenbin Ge and
                  Sibo Song and
                  Kai Dang and
                  Peng Wang and
                  Shijie Wang and
                  Jun Tang and
                  Humen Zhong and
                  Yuanzhi Zhu and
                  Ming{-}Hsuan Yang and
                  Zhaohai Li and
                  Jianqiang Wan and
                  Pengfei Wang and
                  Wei Ding and
                  Zheren Fu and
                  Yiheng Xu and
                  Jiabo Ye and
                  Xi Zhang and
                  Tianbao Xie and
                  Zesen Cheng and
                  Hang Zhang and
                  Zhibo Yang and
                  Haiyang Xu and
                  Junyang Lin},
  title        = {Qwen2.5-VL Technical Report},
  journal      = {CoRR},
  volume       = {abs/2502.13923},
  year         = {2025},
}

@article{wang2024videocot,
  title={Videocot: A video chain-of-thought dataset with active annotation tool},
  author={Wang, Yan and Zeng, Yawen and Zheng, Jingsheng and Xing, Xiaofen and Xu, Jin and Xu, Xiangmin},
  journal={CoRR},
  year={2024}
}

@article{hu2025cos,
  author       = {Jian Hu and
                  Zixu Cheng and
                  Chenyang Si and
                  Wei Li and
                  Shaogang Gong},
  title        = {CoS: Chain-of-Shot Prompting for Long Video Understanding},
  journal      = {CoRR},
  volume       = {abs/2502.06428},
  year         = {2025},
}

@inproceedings{fei2024video,
  author       = {Hao Fei and
                  Shengqiong Wu and
                  Wei Ji and
                  Hanwang Zhang and
                  Meishan Zhang and
                  Mong{-}Li Lee and
                  Wynne Hsu},
  title        = {Video-of-Thought: Step-by-Step Video Reasoning from Perception to
                  Cognition},
  booktitle    = ICML,
  year         = {2024},
}

@inproceedings{yamada2025l3go,
  title={L3go: Language agents with chain-of-3d-thoughts for generating unconventional objects},
  author={Yamada, Yutaro and Chandu, Khyathi and Lin, Bill Yuchen and Hessel, Jack and Yildirim, Ilker and Choi, Yejin},
  booktitle=NAACL,
  pages={456--469},
  year={2025}
}

@inproceedings{li2023intentqa,
  author       = {Jiapeng Li and
                  Ping Wei and
                  Wenjuan Han and
                  Lifeng Fan},
  title        = {IntentQA: Context-aware Video Intent Reasoning},
  booktitle    = ICCV,
  pages        = {11929--11940},
  year         = {2023},
}

@inproceedings{katara2024gen2sim,
  author       = {Pushkal Katara and
                  Zhou Xian and
                  Katerina Fragkiadaki},
  title        = {Gen2Sim: Scaling up Robot Learning in Simulation with Generative Models},
  booktitle    = ICRA,
  pages        = {6672--6679},
  year         = {2024},
}

@article{yuan20243d,
  author       = {Zeqing Yuan and
                  Haoxuan Lan and
                  Qiang Zou and
                  Junbo Zhao},
  title        = {3D-PreMise: Can Large Language Models Generate 3D Shapes with Sharp
                  Features and Parametric Control?},
  journal      = {CoRR},
  volume       = {abs/2401.06437},
  year         = {2024},
}

@inproceedings{wu2024dettoolchain,
  author       = {Yixuan Wu and
                  Yizhou Wang and
                  Shixiang Tang and
                  Wenhao Wu and
                  Tong He and
                  Wanli Ouyang and
                  Philip Torr and
                  Jian Wu},
  title        = {DetToolChain: {A} New Prompting Paradigm to Unleash Detection Ability
                  of {MLLM}},
  booktitle    = ECCV,
  volume       = {15090},
  pages        = {164--182},
  year         = {2024},
}

@inproceedings{mondal2024kam,
  author       = {Debjyoti Mondal and
                  Suraj Modi and
                  Subhadarshi Panda and
                  Rituraj Singh and
                  Godawari Sudhakar Rao},
  title        = {KAM-CoT: Knowledge Augmented Multimodal Chain-of-Thoughts Reasoning},
  booktitle    = AAAI,
  pages        = {18798--18806},
  year         = {2024},
}

@inproceedings{shao2024visual,
  author       = {Hao Shao and
                  Shengju Qian and
                  Han Xiao and
                  Guanglu Song and
                  Zhuofan Zong and
                  Letian Wang and
                  Yu Liu and
                  Hongsheng Li},
  title        = {Visual CoT: Advancing Multi-Modal Language Models with a Comprehensive
                  Dataset and Benchmark for Chain-of-Thought Reasoning},
  booktitle    = NIPS,
  year         = {2024},
}

@inproceedings{xu2024pointllm,
  title={PointLLM: Empowering Large Language Models to Understand Point Clouds},
  author={Xu, Runsen and Wang, Xiaolong and Wang, Tai and Chen, Yilun and Pang, Jiangmiao and Lin, Dahua},
  booktitle={ECCV},
  volume={15083},
  pages={131--147},
  year={2024},
}

@inproceedings{cahyawijaya2024llms,
  author       = {Samuel Cahyawijaya and
                  Holy Lovenia and
                  Pascale Fung},
  title        = {LLMs Are Few-Shot In-Context Low-Resource Language Learners},
  booktitle    = NAACL,
  pages        = {405--433},
  year         = {2024},
}

@inproceedings{gao2025aim,
  author       = {Jun Gao and
                  Qian Qiao and
                  Tianxiang Wu and
                  Zili Wang and
                  Ziqiang Cao and
                  Wenjie Li},
  title        = {{AIM:} Let Any Multimodal Large Language Models Embrace Efficient
                  In-Context Learning},
  booktitle    = AAAI,
  pages        = {3077--3085},
  year         = {2025},
}

@inproceedings{wang2024cog,
  author       = {Shaowei Wang and
                  Lingling Zhang and
                  Longji Zhu and
                  Tao Qin and
                  Kim{-}Hui Yap and
                  Xinyu Zhang and
                  Jun Liu},
  title        = {CoG-DQA: Chain-of-Guiding Learning with Large Language Models for
                  Diagram Question Answering},
  booktitle    = CVPR,
  pages        = {13969--13979},
  year         = {2024},
}

@article{achiam2023gpt,
  author       = {OpenAI},
  title        = {{GPT-4} Technical Report},
  journal      = {CoRR},
  volume       = {abs/2303.08774},
  year         = {2023},
}

@inproceedings{liu2024improved,
  author       = {Haotian Liu and
                  Chunyuan Li and
                  Yuheng Li and
                  Yong Jae Lee},
  title        = {Improved Baselines with Visual Instruction Tuning},
  booktitle    = CVPR,
  pages        = {26286--26296},
  year         = {2024},
}

@inproceedings{liu2023visual,
  author       = {Haotian Liu and
                  Chunyuan Li and
                  Qingyang Wu and
                  Yong Jae Lee},
  title        = {Visual Instruction Tuning},
  booktitle    = NIPS,
  year         = {2023},
}

@inproceedings{wei2022chain,
  author       = {Jason Wei and
                  Xuezhi Wang and
                  Dale Schuurmans and
                  Maarten Bosma and
                  Brian Ichter and
                  Fei Xia and
                  Ed H. Chi and
                  Quoc V. Le and
                  Denny Zhou},
  title        = {Chain-of-Thought Prompting Elicits Reasoning in Large Language Models},
  booktitle    = NIPS,
  year         = {2022},
}

@inproceedings{brown2020language,
  author       = {Tom B. Brown and
                  Benjamin Mann and
                  Nick Ryder and
                  Melanie Subbiah and
                  Jared Kaplan and
                  Prafulla Dhariwal and
                  Arvind Neelakantan and
                  Pranav Shyam and
                  Girish Sastry and
                  Amanda Askell and
                  Sandhini Agarwal and
                  Ariel Herbert{-}Voss and
                  Gretchen Krueger and
                  Tom Henighan and
                  Rewon Child and
                  Aditya Ramesh and
                  Daniel M. Ziegler and
                  Jeffrey Wu and
                  Clemens Winter and
                  Christopher Hesse and
                  Mark Chen and
                  Eric Sigler and
                  Mateusz Litwin and
                  Scott Gray and
                  Benjamin Chess and
                  Jack Clark and
                  Christopher Berner and
                  Sam McCandlish and
                  Alec Radford and
                  Ilya Sutskever and
                  Dario Amodei},
  title        = {Language Models are Few-Shot Learners},
  booktitle    = NIPS,
  year         = {2020},
}

@inproceedings{gao2024cantor,
  author       = {Timin Gao and
                  Peixian Chen and
                  Mengdan Zhang and
                  Chaoyou Fu and
                  Yunhang Shen and
                  Yan Zhang and
                  Shengchuan Zhang and
                  Xiawu Zheng and
                  Xing Sun and
                  Liujuan Cao and
                  Rongrong Ji},
  title        = {Cantor: Inspiring Multimodal Chain-of-Thought of {MLLM}},
  booktitle    = ACMMM,
  pages        = {9096--9105},
  year         = {2024},
}

@inproceedings{li2023blip2,
  author       = {Junnan Li and
                  Dongxu Li and
                  Silvio Savarese and
                  Steven C. H. Hoi},
  title        = {{BLIP-2:} Bootstrapping Language-Image Pre-training with Frozen Image
                  Encoders and Large Language Models},
  booktitle    = ICML,
  volume       = {202},
  pages        = {19730--19742},
  year         = {2023},
}

@article{touvron2023llama,
  author       = {Hugo Touvron and
                  Thibaut Lavril and
                  Gautier Izacard and
                  Xavier Martinet and
                  Marie{-}Anne Lachaux and
                  Timoth{\'{e}}e Lacroix and
                  Baptiste Rozi{\`{e}}re and
                  Naman Goyal and
                  Eric Hambro and
                  Faisal Azhar and
                  Aur{\'{e}}lien Rodriguez and
                  Armand Joulin and
                  Edouard Grave and
                  Guillaume Lample},
  title        = {LLaMA: Open and Efficient Foundation Language Models},
  journal      = {CoRR},
  volume       = {abs/2302.13971},
  year         = {2023},
}

@inproceedings{deitke2023objaverse3,
  author       = {Matt Deitke and
                  Dustin Schwenk and
                  Jordi Salvador and
                  Luca Weihs and
                  Oscar Michel and
                  Eli VanderBilt and
                  Ludwig Schmidt and
                  Kiana Ehsani and
                  Aniruddha Kembhavi and
                  Ali Farhadi},
  title        = {Objaverse: {A} Universe of Annotated 3D Objects},
  booktitle    = CVPR,
  pages        = {13142--13153},
  year         = {2023},
}

@article{team2023gemini,
  author       = {Rohan Anil and
                  Sebastian Borgeaud and
                  Yonghui Wu and
                  Jean{-}Baptiste Alayrac and
                  Jiahui Yu and
                  Radu Soricut and
                  Johan Schalkwyk and
                  Andrew M. Dai and
                  Anja Hauth and
                  Katie Millican and
                  David Silver and
                  Slav Petrov and
                  Melvin Johnson and
                  Ioannis Antonoglou and
                  Julian Schrittwieser and
                  Amelia Glaese and
                  Jilin Chen and
                  Emily Pitler and
                  Timothy P. Lillicrap and
                  Angeliki Lazaridou and
                  Orhan Firat and
                  James Molloy and
                  Michael Isard and
                  Paul Ronald Barham and
                  Tom Hennigan and
                  Benjamin Lee and
                  Fabio Viola and
                  Malcolm Reynolds and
                  Yuanzhong Xu and
                  Ryan Doherty and
                  Eli Collins and
                  Clemens Meyer and
                  Eliza Rutherford and
                  Erica Moreira and
                  Kareem Ayoub and
                  Megha Goel and
                  George Tucker and
                  Enrique Piqueras and
                  Maxim Krikun and
                  Iain Barr and
                  Nikolay Savinov and
                  Ivo Danihelka and
                  Becca Roelofs and
                  Ana{\"{\i}}s White and
                  Anders Andreassen and
                  Tamara von Glehn and
                  Lakshman Yagati and
                  Mehran Kazemi and
                  Lucas Gonzalez and
                  Misha Khalman and
                  Jakub Sygnowski and
                  et al.},
  title        = {Gemini: {A} Family of Highly Capable Multimodal Models},
  journal      = {CoRR},
  volume       = {abs/2312.11805},
  year         = {2023},
}

@inproceedings{gao2021simcse,
  author       = {Tianyu Gao and
                  Xingcheng Yao and
                  Danqi Chen},
  title        = {SimCSE: Simple Contrastive Learning of Sentence Embeddings},
  booktitle    = EMNLP,
  pages        = {6894--6910},
  year         = {2021},
}

@inproceedings{xue2024ulip,
  author       = {Le Xue and
                  Ning Yu and
                  Shu Zhang and
                  Artemis Panagopoulou and
                  Junnan Li and
                  Roberto Mart{\'{\i}}n{-}Mart{\'{\i}}n and
                  Jiajun Wu and
                  Caiming Xiong and
                  Ran Xu and
                  Juan Carlos Niebles and
                  Silvio Savarese},
  title        = {{ULIP-2:} Towards Scalable Multimodal Pre-Training for 3D Understanding},
  booktitle    = CVPR,
  pages        = {27081--27091},
  year         = {2024},
}

@inproceedings{reimers2019sentence,
  author       = {Nils Reimers and
                  Iryna Gurevych},
  title        = {Sentence-BERT: Sentence Embeddings using Siamese BERT-Networks},
  booktitle    = EMNLP_IJCNLP,
  pages        = {3980--3990},
  year         = {2019},
}

@inproceedings{zheng2023ddcot,
  author       = {Ge Zheng and
                  Bin Yang and
                  Jiajin Tang and
                  Hong{-}Yu Zhou and
                  Sibei Yang},
  title        = {DDCoT: Duty-Distinct Chain-of-Thought Prompting for Multimodal Reasoning
                  in Language Models},
  booktitle    = NIPS,
  year         = {2023},
}

@article{hu2025socratic,
  author       = {Wanpeng Hu and
                  Haodi Liu and
                  Lin Chen and
                  Feng Zhou and
                  Changming Xiao and
                  Qi Yang and
                  Changshui Zhang},
  title        = {Socratic Questioning: Learn to Self-guide Multimodal Reasoning in
                  the Wild},
  journal      = {CoRR},
  volume       = {abs/2501.02964},
  year         = {2025},
}

@inproceedings{qi2024shapellm,
  author       = {Zekun Qi and
                  Runpei Dong and
                  Shaochen Zhang and
                  Haoran Geng and
                  Chunrui Han and
                  Zheng Ge and
                  Li Yi and
                  Kaisheng Ma},
  title        = {ShapeLLM: Universal 3D Object Understanding for Embodied Interaction},
  booktitle    = ECCV,
  volume       = {15101},
  pages        = {214--238},
  year         = {2024},
}

@inproceedings{luo2023scalable,
  author       = {Tiange Luo and
                  Chris Rockwell and
                  Honglak Lee and
                  Justin Johnson},
  title        = {Scalable 3D Captioning with Pretrained Models},
  booktitle    = NIPS,
  year         = {2023},
}

\appendix

\end{document}